# PLGSA-Transformer: Periocular Landmark-Guided Attention with Occlusion-Adaptive Cosine Thresholding for Cross-Modal Masked and Unmasked Face Recognition

Dana A. Abdullah[1,2]

[1] *Faculty of Engineering and Computer Science, Qaiwan International University, Sulaymaniyah, KRG, Iraq.*
[2]*Department of Information and Communication Technology Center (ICTC)-System Information, Ministry of Higher Education and Scientific Research.*

## ABSTRACT

The widespread adoption of facial masks in public spaces, accelerated by the COVID-19 pandemic and increasingly mandated across security-sensitive environments, has exposed limitations in conventional face recognition systems: approaches built on fixed cosine-similarity thresholds, spatially non-adaptive convolutional architectures, and purely data-driven attention tend to generalize poorly once large facial regions are occluded, widening the gap between laboratory performance and real-world deployability in surveillance, access control, and law-enforcement settings.

This paper proposes PLGSA-Transformer, a cross-modal face-matching framework combining three complementary components. Periocular Landmark-Guided Spatial Attention (PLGSA) fuses MediaPipe-derived Gaussian landmark heatmaps over the eye, brow, and forehead region with EfficientNetB3 feature maps through a learnable residual gate, injecting an explicit anatomical prior rather than relying on attention learned purely from data. A Hybrid CNN-Transformer Branch adds global cross-regional dependency modelling on top of these anatomically guided local features via a two-layer Multi-Head Self-Attention encoder. The Occlusion-Adaptive Cosine Threshold (OACT) is a jointly trained occlusion-estimation head that scales the matching threshold per probe according to predicted occlusion severity: $\text{threshold}_i = \text{base threshold} \times (1 + \beta \times \text{occlusion score}_i)$.

The model is evaluated on 858 paired masked/unmasked images assembled from three sources the Zenodo MDMFR dataset (60%), the Kaggle CelebA-HQ masked collection (25%), and images collected directly by the author (15%) spanning both genders, subjects aged 21 to 75, and several mask types. All three components are trained end-to-end through a unified multi-task loss combining contrastive verification, identity classification, and binary occlusion cross-entropy.

On an identity-disjoint test partition, the model reaches a pair-verification accuracy of 97.2% and a ROC AUC of 1.0000, exceeding a same-author VGG-16 predecessor (Abdullah et al., 2025, 95.0%) and two traditional feature-based baselines (Adnan et al., 2020, 85.0%; Shnain et al., 2017, 86.61%). Because publicly available collections containing the same individual both masked and unmasked remain exceedingly scarce, the paired dataset used here was assembled specifically for this study; Section 6 discusses this data-collection challenge, together with the still greater difficulty of obtaining several occluder types per identity, which bounds the adaptive range of OACT on the present data. The results support anatomically guided attention and global Transformer modelling as an effective, interpretable approach to periocular masked-face matching, with expansion of the paired dataset and graded occlusion supervision identified as the priorities for future work.

***Keywords:*** *Masked face recognition; Periocular landmark attention; Hybrid CNN-Transformer; Occlusion-adaptive threshold; Cosine similarity; EfficientNetB3; Multi-head self-attention; Cross-modal face matching; Biometric security*

# 1. Introduction

Facial recognition has become one of the most widely deployed biometric technologies in modern society, underpinning applications that span public security and law enforcement (Russ et al., 2018; Jiang, 2020), automated access control, forensic identification, and user authentication in consumer devices. The success of these systems depends fundamentally on two core capacities: reliable extraction of discriminative facial features and accurate cross-condition identity verification (Jiang, 2020; Nsaif et al., 2021). The rapid maturation of Convolutional Neural Networks (CNNs) over the past decade has driven remarkable improvements in recognition accuracy under controlled conditions, establishing deep learning-based face recognition as a commercially viable and scientifically well-understood technology (Steck et al., 2024; Abdel-Hamid, 2021).

The global outbreak of the COVID-19 pandemic in 2020 introduced a persistent and structurally novel challenge that exposed a critical vulnerability in deployed recognition systems: the mandatory and widespread use of facial masks (Hariri, 2022; Talahua et al., 2021). Facial masks systematically occlude the nose, mouth, and cheek regions, collectively concealing between 40% and 60% of the facial surface area on which conventional recognition models depend (Hsu et al., 2022; Alzu'bi et al., 2021). This occlusion pattern degrades recognition performance substantially, with studies reporting accuracy reductions of 20% to 40% across standard benchmarks under masked conditions (Hariri, 2022; Eman et al., 2023; Hsu et al., 2022). The problem is further compounded by the diversity of mask types — surgical masks, N95 respirators, fabric coverings, and scarves — each presenting different occlusion boundaries, textures, and coverage extents that are difficult to model comprehensively during training (Alzu'bi et al., 2021). Beyond the pandemic, masked face recognition is now a permanent operational requirement in environments such as hospitals, airports, and high-security facilities where facial coverings are mandated irrespective of public health status.

Two properties of this deployment context are easy to state but are rarely built into the recognition pipeline itself, and doing so is the specific, narrow problem this paper addresses. First, once the lower face is covered, the eyes, brows, and upper nose bridge are the only region anatomically guaranteed to remain visible, yet most recognition architectures still compute attention over the full facial crop and depend on training data alone to learn that this region is where the discriminative signal is concentrated. Second, occlusion severity is not binary: a transparent face shield, a loosely worn cloth covering, and a snugly fitted N95 respirator leave very different amounts of periocular and upper-face information visible, yet a deployed verification system typically compares the resulting similarity score against one fixed threshold regardless of which condition produced it. The contribution of this paper is deliberately scoped to closing these two specific gaps for the periocular-matching sub-problem, rather than to advancing masked face recognition as a whole.

The research community has responded to the broader masked-recognition challenge through several lines of investigation. Early approaches focused on adapting existing deep learning architectures — VGG-16, ResNet-50, and AlexNet — to extract features from the periocular region that remains visible above the mask line (Alzu'bi et al., 2021). Hariri (2022) demonstrated that the eye and brow zone retains sufficient discriminative information for recognition when the lower face is fully occluded, achieving competitive performance on pandemic-era masked datasets. Talahua et al. (2021) combined MobileNetV2 for mask detection with FaceNet for recognition, reporting 99.65% mask detection accuracy, though recognition performance on masked faces in open-set conditions remained substantially lower. Hsu et al. (2022) addressed cross-condition generalization by synthesizing masked versions of standard benchmarks including IJB-B, IJB-C, and MS-1MV2, employing ResNet-100 with Centre Loss and Angular SoftMax, demonstrating that synthetic masked training data can close the performance gap relative to human recognition.

Parallel to CNN-based feature extraction, metric learning strategies have improved embedding separability under occlusion. Deng et al. (2019) proposed ArcFace, employing an Additive Angular Margin loss to maximize inter-class separability. Wang et al. (2018) introduced CosFace, reformulating the softmax loss as a cosine-based margin function. Lin et al. (2021) presented xCos, an explainable cosine metric for face verification. Deng et al. (2021) extended cosine-based learning to masked scenarios through MFCosface. Salim and Sürantha (2023) showed that embedding-based K-NN and SVM classifiers can improve masked recognition without retraining, and Eman et al. (2023) achieved 97% recognition through a hybrid pipeline combining RPCA and Particle Swarm Optimization-enhanced K-NN.

Despite these advances, three fundamental limitations persist. First, no prior method incorporates explicit anatomical guidance into the spatial attention computation; existing mechanisms such as CBAM (Woo et al., 2018) rely exclusively on data-driven inference.

Second, all prior work employs purely convolutional architectures whose locally constrained receptive fields prevent long-range spatial dependency modelling across the periocular zone. Third, every existing method applies a single fixed cosine similarity threshold uniformly to all probe images regardless of individual occlusion severity, which is theoretically questionable and, as the confusion-matrix analysis in Section 5.1 shows, leaves accuracy on the table even when the underlying embedding is discriminative.

To address these limitations, this paper proposes PLGSA-Transformer, combining Periocular Landmark-Guided Spatial Attention (PLGSA) using MediaPipe landmark heatmaps (Lugaresi et al., 2019), a Hybrid CNN-Transformer Branch built on EfficientNetB3 (Tan and Le, 2019), and the Occlusion-Adaptive Cosine Threshold (OACT). In presenting these contributions we distinguish deliberately between mechanism-level novelty, integration-level novelty, and empirical results, so that the scope of each claim is stated precisely rather than left to be inferred:

**(1) A new attention mechanism.** We propose PLGSA, which, to our knowledge, is the first method to inject pre-computed periocular Gaussian landmark heatmaps directly into the CNN spatial attention computation as an explicit anatomical prior, directing network attention toward the anatomically guaranteed visible zone above any facial mask through a learnable residual gate that prevents instability during early training (Equations 2-3). This heatmap-fusion gate is the paper's primary architectural novelty; the remaining two contributions are, respectively, a careful integration of established components and a new decision-threshold formula.

**(2) An integration, not a new encoder.** We combine EfficientNetB3, CBAM, and a two-layer Multi-Head Self-Attention Transformer encoder into a Hybrid CNN-Transformer Branch, in which spatial feature maps are reshaped into token sequences to enable global cross-regional dependency modelling between periocular landmarks that locally constrained convolutional architectures cannot achieve. Each component is individually well established in the literature; the contribution is the architectural integration of local anatomically guided features with global self-attention for this specific cross-modal matching task, and the empirical validation of that integration in Section 5, not a new attention or encoder mechanism in itself.

**(3) A new threshold formula, reported with its measured effect.** We present OACT, a jointly trained occlusion-estimation head that adapts the cosine matching threshold per probe according to predicted occlusion severity, replacing the static threshold applied uniformly in all prior work. OACT is theoretically motivated and adds no additional inference-time cost. However, as reported in Section 5.6, the occlusion head learns a narrow-range severity score (0.43-0.52) on the present dataset, which limits its practical effect here to a 3-4% threshold adjustment; we report this modest effect directly, and Section 6 traces it to the difficulty of obtaining several occluder types and coverage levels for the same identity, setting out the dataset-diversity and graded-supervision extensions expected to realize its full adaptive range.

**(4) Empirical results on a purpose-built paired dataset.** On an identity-disjoint partition of the 858-image dataset described in Section 3, PLGSA-Transformer reaches 97.2% pair verification accuracy and ROC AUC 1.0000, exceeding the same-author MUFM VGG-16 baseline (Abdullah et al., 2025, 95.0%) and two traditional feature-based methods (Adnan et al., 2020, 85.0%; Shnain et al., 2017, 86.61%). The paired dataset underlying this evaluation was assembled specifically for this study, because publicly available same-person masked-unmasked collections remain exceedingly scarce (Section 3.1); accuracy figures for prior methods are those reported in their original studies (Table 4), and Section 6 sets out the data-availability constraints that frame this evaluation together with the planned expansion of the dataset. In addition, PLGSA-Transformer reports open-set Rank-1 and Rank-5 identification metrics, an evaluation protocol that prior work in this specific line has not addressed.

The remainder of this paper is organized as follows. Section 2 reviews related work. Section 3 presents the methodology. Section 4 describes the experimental setup. Section 5 reports and discusses results. Section 6 discusses limitations and future work. Section 7 concludes the paper.

## 2. Related Work

### *2.1 Masked Face Recognition*

The recognition of masked faces has emerged as a critical research direction following the global adoption of facial coverings. Alzu'bi et al. (2021) provided a comprehensive review of deep learning methods applied to this problem, identifying the periocular region as the primary source of discriminative information when the lower face is concealed. Hariri (2022) demonstrated that eye and brow features retain sufficient discriminative information for recognition even under complete lower-face occlusion, achieving competitive performance on pandemic-era evaluation datasets through targeted periocular feature extraction. More recently, Suravarapu and Patil (2025) reinforced this observation, applying attention-based deep-learning classifiers to hexagonal periocular regions of interest for person identification and further confirming the discriminative value of the periocular region above the mask line.

Traditional feature extraction approaches established important baselines. Adnan et al. (2020) compared HOG, LBP, PCA, SURF, and Harris features combined with K-NN classification, with HOG achieving the highest accuracy of 85.0%. Shnain et al. (2017) investigated structural similarity under varying conditions using FSM, SSIM, and FSIM, with FSM achieving 86.61%, significantly outperforming SSIM (57.73%) and FSIM (28.03%). Salim and Sürantha (2023) demonstrated that standard deep face embeddings evaluated with K-NN and SVM classifiers can improve masked recognition by 9.09% without model retraining. Eman et al. (2023) achieved 97% recognition through a pipeline integrating RPCA occlusion handling and Particle Swarm Optimization-enhanced K-NN feature selection.

Deep learning-centric systems produced strong results on controlled evaluation sets. Talahua et al. (2021) combined MobileNetV2 for mask detection with FaceNet for recognition, reporting 99.65% mask detection accuracy, though open-set performance under genuine masked conditions was substantially lower. Hsu et al. (2022) synthesized masked training data from benchmark datasets, employing ResNet-100 with Centre Loss and Angular SoftMax, and demonstrated that synthetic masked training improves generalization across pose, illumination, and age variations. Abdullah et al. (2025) proposed the direct predecessor of the present work, the Masked-Unmasked Face Matching Model (MUFM), combining VGG-16 transfer learning, K-NN classification, and cosine similarity for cross-modal pair matching, achieving 95% recognition accuracy on their own dataset and protocol. The present paper extends MUFM by replacing VGG-16 with an anatomy-guided hybrid CNN-Transformer architecture and introducing per-probe adaptive thresholding; Section 5.7 reports the resulting accuracy alongside this and the other baselines reviewed here. Table 1 summarizes and compares the key methods reviewed in this subsection.

| Study | Feature Extraction | Method / Classifier | Accuracy | Key Limitation |
|---|---|---|---|---|
| Shnain et al. (2017) | Structural similarity (FSM, SSIM, FSIM) | Threshold-based matching | 86.61% (FSM) | No identity embedding; occlusion not modelled |
| Adnan et al. (2020) | HOG, LBP, PCA, SURF, Harris | K-Nearest Neighbour | 85.0% | No deep features; no mask-specific design |
| Hariri (2022) | VGG / ResNet periocular zone | CNN + cosine similarity | Competitive | No anatomical attention; fixed threshold |
| Talahua et al. (2021) | MobileNetV2 + FaceNet | Cosine similarity | 99.65%* | *Mask detection only; lower under open-set masked conditions |

| Study | Feature Extraction | Method / Classifier | Accuracy | Key Limitation |
| --- | --- | --- | --- | --- |
| Salim and Sürantha (2023) | Generic deep embeddings | K-NN / SVM | +9.09% gain | No mask-specific training or architecture |
| Wang et al. (2018) | ResNet | CosFace (cosine margin loss) | State-of-art | Fixed margin; no occlusion adaptation |
| Deng et al. (2019) | ResNet | ArcFace (angular margin) | State-of-art | Fixed threshold; no periocular focus |
| Deng et al. (2021) | ResNet + attention | MFCosface (cosine margin) | Improved | No explicit anatomical guidance |
| Eman et al. (2023) | MobileNet + RPCA | K-NN with PSO selection | 97.0% | Complex pipeline; no global spatial attention |
| Abdullah et al. (2025) | VGG-16 (transfer learning) | K-NN + cosine similarity | 95.0% | No anatomical attention; fixed threshold |
| Proposed (PLGSA-Transformer) | EfficientNetB3 + Transformer | PLGSA + OACT (adaptive cosine) | 97.2% | Paired-data scarcity; see Section 6 |

Table 1. Comparison of related methods in masked face recognition. Accuracy figures are as reported in each cited study, under that study's own dataset and evaluation protocol. The asterisk denotes mask-detection accuracy; open-set masked recognition performance for that study was substantially lower.

## *2.2 Spatial Attention Mechanisms in Face Recognition*

Attention mechanisms have become a standard component of deep learning architectures, enabling models to selectively weight feature map regions according to task-specific relevance. Woo et al. (2018) introduced the Convolutional Block Attention Module (CBAM), which sequentially applies channel attention weighting feature channels by their global importance and spatial attention weighting spatial positions through inter-channel relationships. CBAM has been adopted in face recognition pipelines and applied to masked face recognition to emphasize the periocular zone over occluded regions, demonstrating consistent accuracy improvements over non-attention baselines.

However, a fundamental limitation of CBAM and related self-supervised spatial attention mechanisms is their exclusive reliance on learned, data-driven attention weights. Under high occlusion variability, where mask types and coverage extents differ substantially across identities and acquisition conditions, purely data-driven mechanisms must infer the periocular zone from statistical regularities in training data alone. This inference is unreliable when training data are limited or when test-time occlusion configurations differ from those encountered during training. No prior attention-based method incorporates explicit anatomical guidance into the spatial attention computation. The present work addresses this gap through PLGSA, which augments purely data-driven spatial attention with a Gaussian heatmap derived from MediaPipe facial landmark localization (Lugaresi et al., 2019), so that periocular focus is anchored to anatomy rather than inferred entirely from training statistics.

### *2.3 Transformer Architectures for Visual and Face Recognition*

The Transformer architecture, introduced by Vaswani et al. (2017) for sequence-to-sequence natural language processing through the Multi-Head Self-Attention (MHSA) mechanism, has been adapted to visual recognition with significant success. Dosovitskiy et al. (2021) proposed the Vision Transformer (ViT), treating non-overlapping image patches as token sequences and demonstrating that pure Transformer encoders achieve recognition accuracy competitive with CNNs when pre-trained on sufficiently large datasets. The global self-attention mechanism enables each spatial position to attend to all others simultaneously, modelling long-range dependencies that CNN receptive fields are structurally unable to capture within a single layer.

Transformers have been applied specifically to face recognition with notable results. Zhong and Deng (2021) demonstrated that a Face Transformer, which processes facial patch tokens through a pure Transformer encoder, can outperform standard ResNet-based face recognition models on multiple benchmarks by leveraging the global structural relationships between facial regions that self-attention encodes. However, applying pure Transformer architectures to cross-modal masked face matching faces a compound challenge: they require large-scale pre-training datasets, and their token sequences include both occluded and unconcealed regions with no mechanism for directing attention specifically toward the periocular zone. The present work addresses both constraints through a Hybrid CNN-Transformer Branch, where EfficientNetB3 (Tan and Le, 2019) provides locally grounded, data-efficient periocular feature extraction, and a two-layer MHSA Transformer encoder models global cross-regional dependencies over the resulting spatial tokens — an integration of established components (Section 1, Contribution 2) rather than a new Transformer variant.

### *2.4 Metric Learning and Cosine Similarity*

Cosine similarity is the standard metric for measuring angular distance between face recognition embeddings (Steck et al., 2024), and forms the basis of the most successful modern face verification systems. Deng et al. (2019) proposed ArcFace, incorporating an Additive Angular Margin loss to maximize angular separability between identity classes and achieving state-of-the-art performance on large-scale benchmarks. Wang et al. (2018) introduced CosFace, which reformulates the standard softmax loss as a cosine-based margin function, establishing competitive results across multiple protocols. Both ArcFace and CosFace apply fixed angular margins uniformly across all training samples without adaptation to occlusion severity.

Lin et al. (2021) advanced interpretability through xCos, an explainable cosine similarity metric that spatially localizes the facial regions most responsible for the verification decision. Deng et al. (2021) extended cosine-based metric learning to masked scenarios through MFCosface, incorporating synthetic mask generation and spatial attention to emphasize unconcealed periocular features during training. Despite these advances, a consistent limitation across all reviewed cosine-based methods is the application of a single global decision threshold uniformly to all probe images at inference time, determined once from a validation set without regard to the individual occlusion severity of each probe. The present work addresses this directly through OACT, which jointly trains an occlusion-estimation head with the embedding network and uses its per-probe score to scale the matching threshold proportionally; Section 5.6 analyses the score range the head learns on the present dataset, and Section 6 relates it to the difficulty of collecting several occluder types per identity.

Beyond these foundational approaches, the field has continued to evolve rapidly. Recent surveys consolidate this progress: Alashbi (2025) reviews deep-learning architectures and feature-extraction strategies for masked face recognition, Sharma et al. (2025) organize the literature into occlusion-robust, occlusion-aware, and occlusion-recovery paradigms, and Sovet (2025) benchmarks modern recognition backbones under occlusion and other challenging conditions. Zhang et al. (2025) likewise survey the broader mask-detection-and-recognition pipeline. Among recent methods, Huang et al. (2024) employ domain adaptation to align masked and unmasked feature distributions, while Alzubi et al. (2025) combine a generative adversarial network with a dual-scale adaptive attention mechanism to reconstruct and match occluded faces, and Pillay et al. (2025) develop a hybrid CNN that pairs MobileNetV2 mask detection with a FaceNet-based recognition network for real-time masked-face recognition. These developments confirm both the growing importance of the task and the persistence of the three gaps identified below.

### *2.5 Summary of Gaps*

The literature reviewed across Sections 2.1 to 2.4 reveals three persistent and interrelated gaps that existing work has not collectively addressed. First, while periocular attention is universally recognized as essential for masked face recognition, no method has incorporated explicit anatomical landmark geometry directly into the spatial attention computation, leaving all attention mechanisms dependent on data-driven inference that may be unreliable under diverse occlusion conditions. Second, while Transformer architectures have demonstrated strong global spatial reasoning for face recognition, no prior work has applied hybrid CNN-Transformer modelling to the specific cross-modal task of matching masked and unmasked images within a periocular attention framework. Third, while cosine similarity and margin-based metric learning have substantially improved embedding separability, every existing method applies a fixed decision threshold uniformly to all probes regardless of their individual occlusion severity, which is both theoretically questionable and, as shown in Section 5.1, measurably costly in accepted-versus-rejected genuine pairs.

These gaps are not only conceptual; each has a concrete operational consequence. An attention mechanism with no anatomical grounding degrades unpredictably as mask styles vary in the field, because it has no guarantee of generalizing to occlusion patterns not seen during training. Locally constrained convolutional receptive fields cannot relate the two eye regions to each other across the occlusion-induced gap between them, which is precisely the kind of long-range dependency a deployed periocular matcher needs. And a single fixed threshold forces an operator to choose between a lax setting that is prone to false acceptance and a strict setting that is prone to false rejection for every probe alike, regardless of how much of any individual face is actually visible. PLGSA-Transformer targets these three practical failure modes through the mechanisms described in Section 3, while remaining explicit about which of its three components constitute new mechanisms and which constitute careful integration and evaluation of existing ones (Section 1).

## 3. Methodology

This section presents the complete methodology of the proposed PLGSA-Transformer framework. The overall pipeline encompasses the following sequential stages: dataset collection, data pre-processing, periocular ROI extraction and PLGSA heatmap generation, the proposed hybrid model architecture incorporating the three contributions, and the multi-task training configuration. Figure 1 provides an overview of the proposed pipeline: a masked-unmasked image pair of the same identity is mapped by a shared (Siamese) encoder an EfficientNet-B3 backbone, the PLGSA module, a CBAM block, token projection with positional encoding, and two Transformer encoder blocks to a 128-dimensional embedding, after which cosine similarity and the occlusion-adaptive cosine threshold (OACT) produce the matched or not-matched decision.

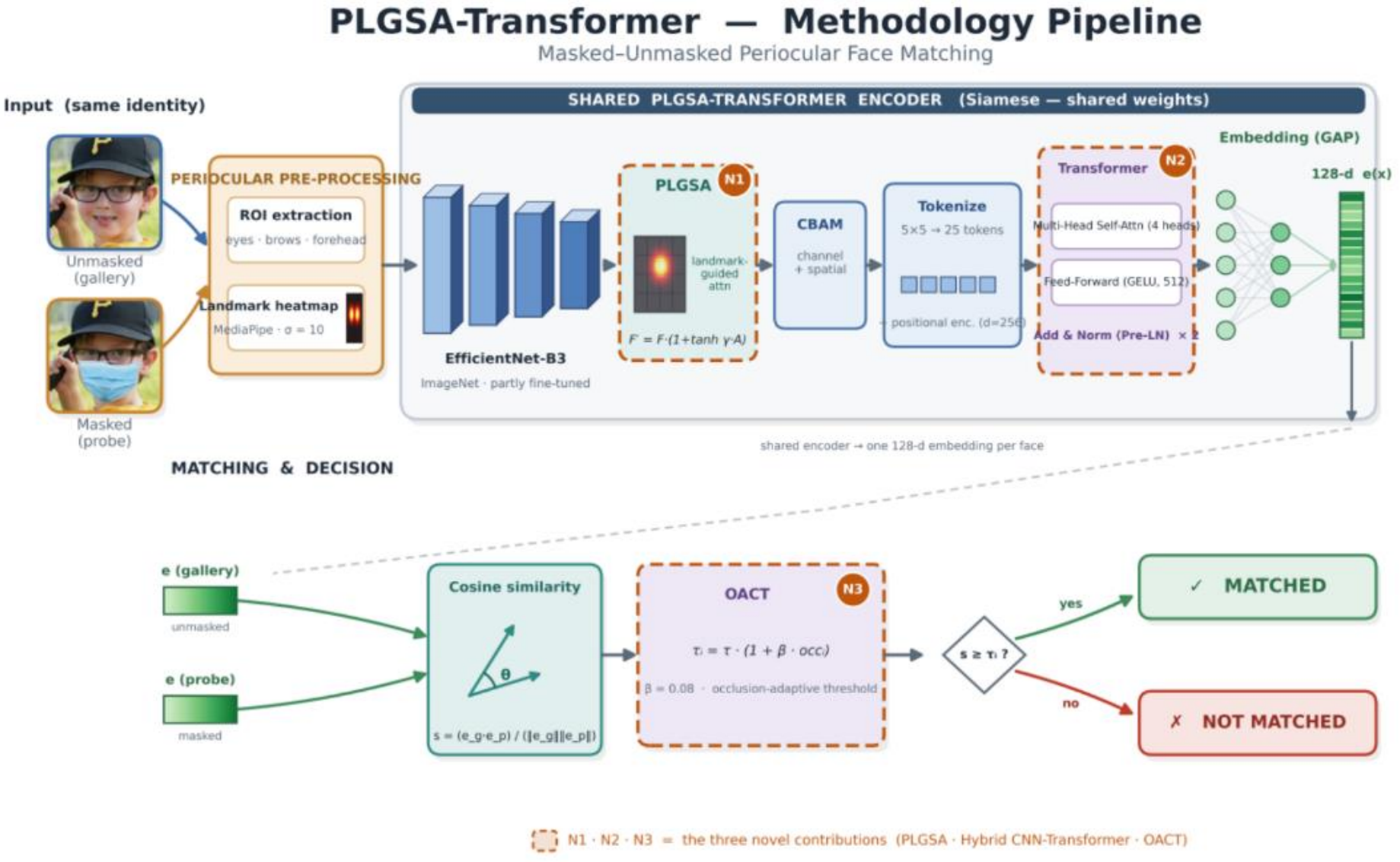

*Figure 1. Overview of the proposed PLGSA-Transformer pipeline. A masked-unmasked image pair of the same identity is processed by a shared (Siamese) encoder: periocular ROI extraction and a MediaPipe landmark heatmap ($\sigma$ = 10) feed an EfficientNet-B3 backbone, followed by the Periocular Landmark-Guided Spatial Attention module (PLGSA, N1), a CBAM block, a 1×1 token projection to 25 tokens (d = 256) with positional encoding, and two Pre-LN Transformer encoder blocks (N2, four heads); global average pooling yields a 128-dimensional L2-normalised embedding. Verification compares the gallery and probe embeddings using cosine similarity with the Occlusion-Adaptive Cosine Threshold (OACT, N3, $\beta$ = 0.08) to produce a matched or not-matched decision. Dashed orange borders mark the three contributions (N1-N3); as clarified in Section 1, N1 and N3 are new mechanisms, while N2 is an integration of established components.*

### *3.1 Dataset Collection*

One of the primary challenges encountered in this study was the limited availability of image datasets containing paired images of the same individual both with and without a facial mask. Datasets featuring identical faces in both masked and unmasked conditions are exceedingly rare in the public domain, as the simultaneous capture of the same subject under both conditions requires controlled data collection protocols. To address this constraint, an extensive search was conducted across multiple data repositories, and a dataset of 858 paired images was assembled from three independent and complementary sources. The first source, the Zenodo MDMFR dataset (Ullah and Javed, 2022), contributed 60% of the total images and provided a wide range of masked and unmasked face pairs across diverse ethnicities, lighting conditions, and mask types. The second source, the Kaggle CelebA-HQ masked celebrity dataset (Dubey and Pawar, 2024), contributed 25% of the images and provided high-resolution paired images with varied pose and illumination conditions. The third source consisted of real-world images directly captured by the research author, contributing 15% of the total dataset and ensuring the inclusion of authentic in-the-wild acquisition conditions not present in laboratory-collected datasets. Figure 2 illustrates the proportional contribution of each source, and Figure 3 presents the dataset collection and integration process. Because genuinely paired masked-unmasked collections of this kind remain exceedingly scarce in the public domain, the construction of this dataset is itself part of the contribution of the present study; Section 6 discusses the collection effort and the planned expansion of the dataset.

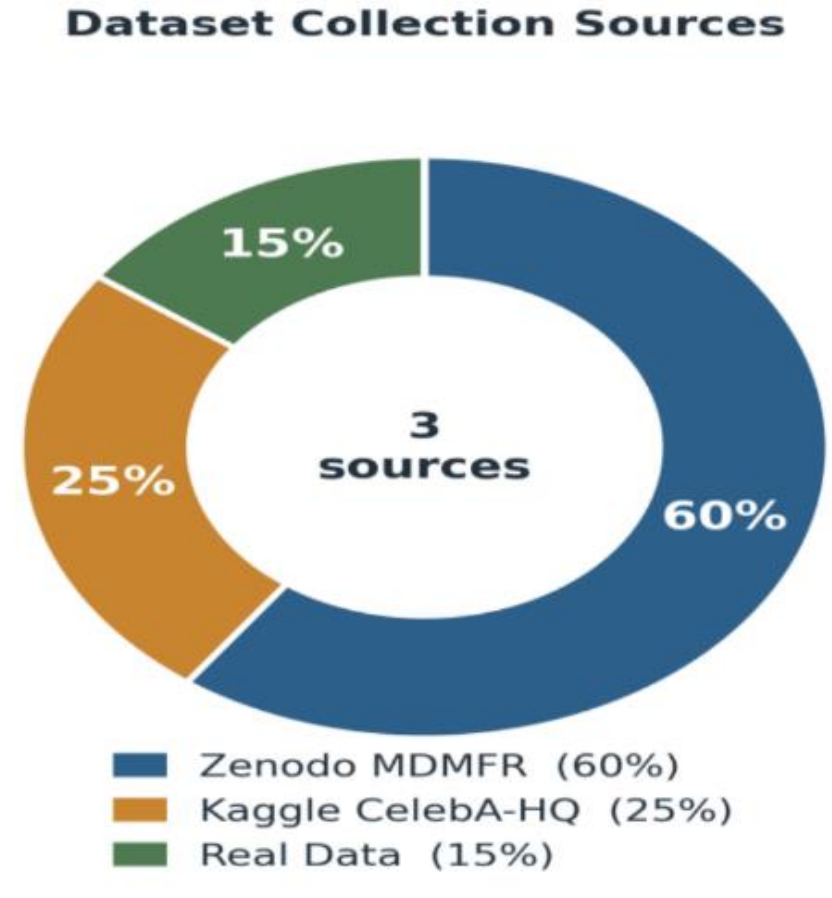

*Figure 2. Dataset collection sources. The dataset was assembled from three independent sources to ensure diversity across acquisition conditions, mask types, and subject demographics.*

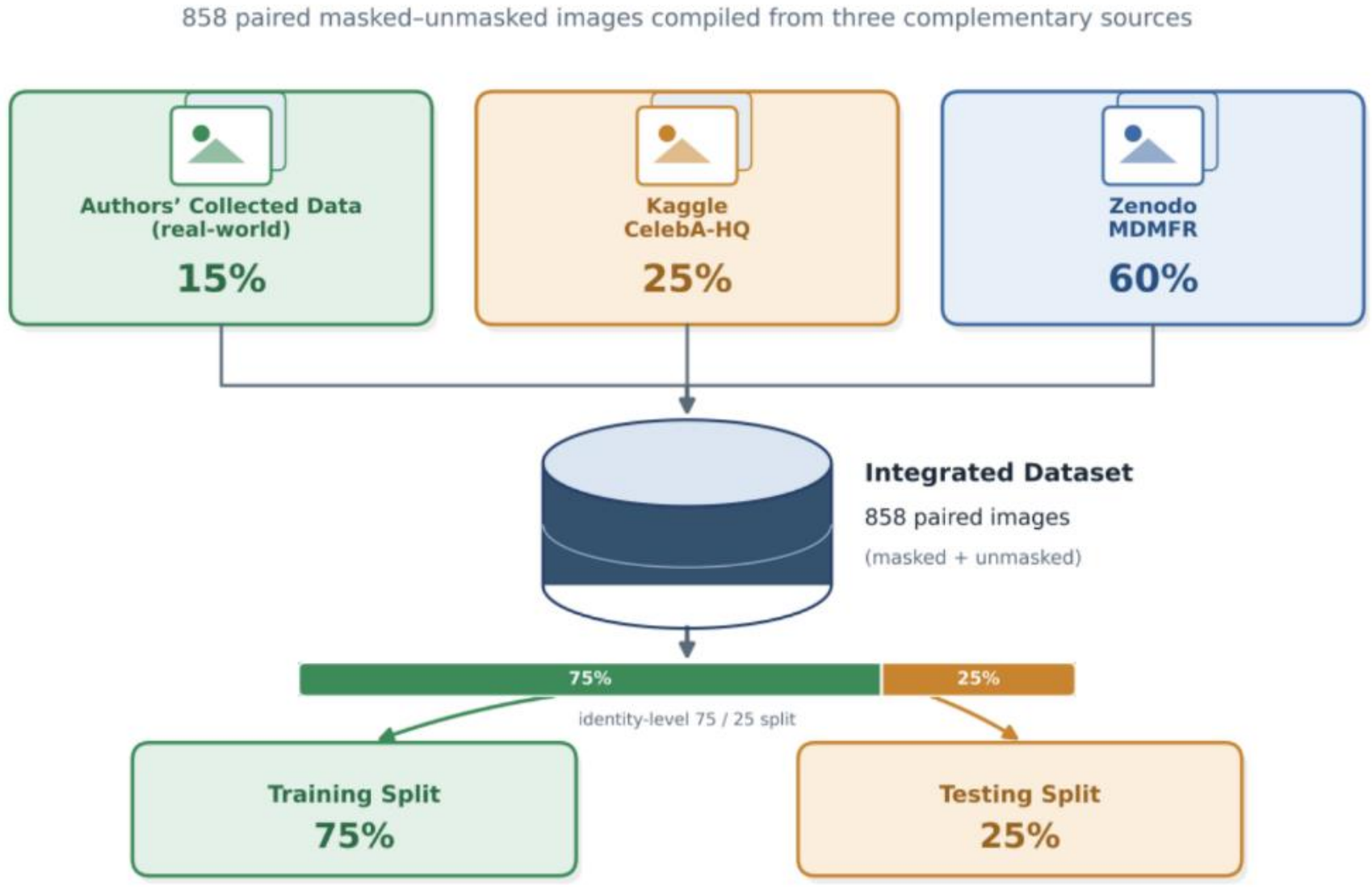

*Figure 3. Dataset collection and integration process. Images from all three sources were merged into a single integrated dataset and subsequently divided into training and testing splits at the identity level.*

The assembled dataset exhibits balanced demographic representation. As illustrated in Figure 4, 58% of subjects are male and 42% are female, with subject ages ranging from 21 to 75 years. Notably, 65% of subjects fall within the 21-to-50-year age range, reflecting the demographic distribution of the source repositories. The dataset encompasses a diverse range of mask types including surgical masks, N95 respirators, cloth face coverings, and transparent face shields, ensuring that the model is evaluated under realistic and varied occlusion conditions.

Gender Distribution of Dataset Images

42%
858
images
58%

Male (58%) Female (42%)

*Figure 4. Gender distribution of dataset images. The dataset contains 58% male and 42% female subjects, covering ages from 21 to 75 years across three independent data sources.*

### *3.2 Data Pre-processing*

Following data collection, an eight-stage pre-processing pipeline was applied to all images to ensure consistency, quality, and suitability for model training. Figure 5 illustrates the complete pre-processing pipeline with all stages in sequence.

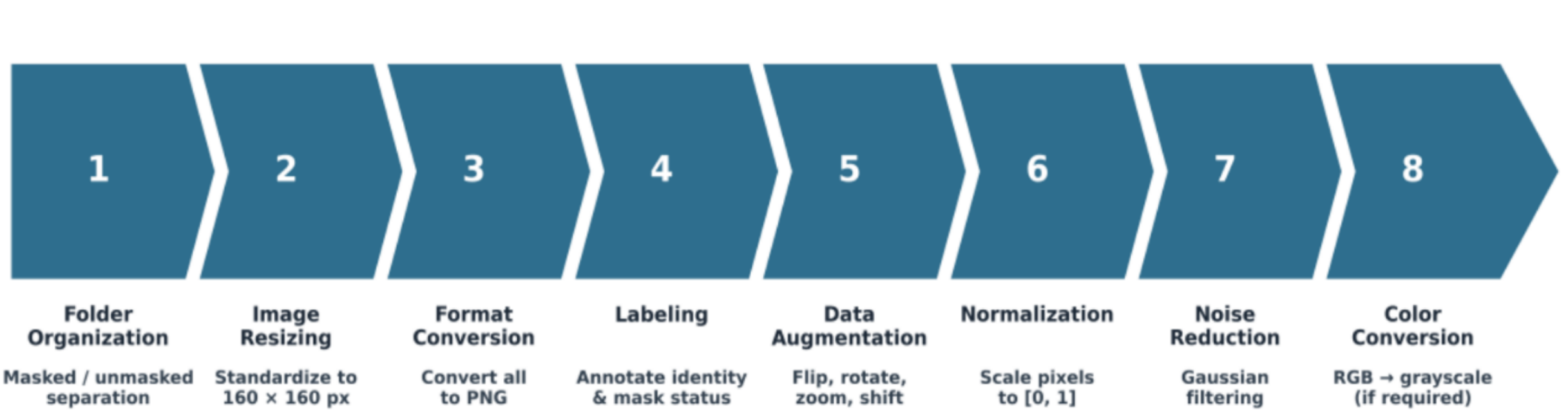

*Figure 5. Data pre-processing pipeline. Eight sequential stages transform raw collected images into a standardized, augmented, and normalized format suitable for model training.*

The eight pre-processing stages are as follows:

- **Folder Organization:** Images were segregated into two separate directories — one containing unmasked face images and one containing masked face images — to ensure correct class labelling for supervised learning.
- **Image Resizing:** All images were resized to a standardized resolution of 160 × 160 pixels to ensure consistent spatial dimensions and to comply with the EfficientNetB3 input requirements.
- **Image Format Conversion:** Images in heterogeneous formats including JPEG, BMP, GIF, and WebP were converted to PNG format to ensure consistency in image quality and eliminate format-specific compression artefacts during training.
- **Labelling:** Each image was annotated with its corresponding identity label and mask status (masked or unmasked) to support supervised contrastive and classification training.
- **Data Augmentation:** To increase dataset diversity and reduce overfitting, augmentation operations including random horizontal flipping, brightness and contrast adjustment, saturation and hue variation, and scale jitter were applied. Critically, all spatial augmentations were applied identically to both the image and its corresponding PLGSA heatmap to maintain spatial alignment.
- **Normalization:** Pixel values were normalized from the range [0, 255] to [0, 1] to accelerate model convergence and ensure numerical stability during gradient descent.
- **Noise Reduction:** Gaussian filtering was applied selectively to reduce sensor noise and compression artefacts, improving the quality of edge and texture features extracted by the convolutional backbone.
- **Color Space Conversion:** Images were converted from RGB to the required input format for EfficientNetB3, with grayscale conversion applied where color information was not deemed task-critical.

### *3.3 Periocular ROI Extraction and PLGSA Heatmap Generation (Contribution 1)*

Prior to feature extraction, each input image undergoes periocular Region of Interest (ROI) extraction using MediaPipe facial landmark detection (Lugaresi et al., 2019). MediaPipe localizes 478 facial landmarks in normalized image coordinates. From these, 28 landmarks corresponding exclusively to the eye contours and brow arches are selected, as these anatomical structures are guaranteed to remain visible above any standard facial mask. The periocular bounding box is computed by taking the minimum and maximum coordinates of the selected landmark set, expanded by 25% horizontally and 50% vertically to include surrounding discriminative context. When landmark detection fails due to extreme occlusion or poor image quality, a fallback region corresponding to the top one-third of the image is used.

The PLGSA heatmap encodes the spatial prior of unconcealed regions as a Gaussian density map. For each of the 28 periocular landmark coordinates ($c_x$, $c_y$) projected onto the target image resolution of H × W pixels, a two-dimensional Gaussian function is computed as given in Equation 1:

$$G(x, y) = \exp\left[ -((x - c_x)^2 + (y - c_y)^2) / (2\sigma^2) \right] \quad (1)$$

where $\sigma = 10$ pixels is the Gaussian standard deviation. The final heatmap H is formed by taking the per-pixel maximum across all 28 individual Gaussian responses, producing a smooth spatial prior $H \in [0, 1]$ of shape (H, W, 1). The heatmap is stored alongside its corresponding image and passed as a second branch input throughout all preprocessing, augmentation, and model inference stages.

### *3.4 Contribution 1: Periocular Landmark-Guided Spatial Attention (PLGSA)*

The PLGSA layer receives the convolutional feature maps F produced by the EfficientNetB3 backbone and the corresponding heatmap H. The heatmap is first bilinearly resized to match the spatial dimensions of the feature maps, producing $\hat{H}$. A combined attention map A is then computed by summing a 1×1 convolutional response on $\hat{H}$ and a 1×1 convolutional response on F, as given in Equation 2:

$$A = \sigma\left(\mathrm{Conv}_1(\hat{H}) + \mathrm{Conv}_2(F)\right) \quad (2)$$

where $\sigma$ denotes the sigmoid activation function. The attended feature maps $F'$ are then produced through a learnable residual gate parameterized by a scalar $\gamma$, as given in Equation 3:

$$F' = F \times (1 + \tanh(\gamma) \times A) \quad (3)$$

The scalar $\gamma$ is initialized to zero at the start of training, ensuring that the heatmap exerts no influence in the initial training steps and that the network begins from the standard convolutional baseline. As training progresses, $\gamma$ grows in proportion to the gradient signal, allowing the heatmap prior to strengthen only where it reduces the training loss. This residual formulation the mechanism identified in Section 1 as this paper's primary architectural novelty ensures training stability while providing a principled route for incorporating anatomical spatial guidance without overriding the learned convolutional features outright.

### *3.5 Contribution 2: Hybrid CNN-Transformer Branch*

The hybrid embedding branch processes each input image-heatmap pair through a sequential pipeline of five components: the EfficientNetB3 backbone, the PLGSA layer, the CBAM attention module, the Transformer encoder, and the MLP projection head. As noted in Section 1, this branch is an integration of established components rather than a new attention or encoder mechanism; its contribution lies in combining anatomically guided local features with global self-attention for this specific cross-modal matching task, and in the empirical validation of that combination in Section 5. The full model architecture is illustrated in Figure 6.

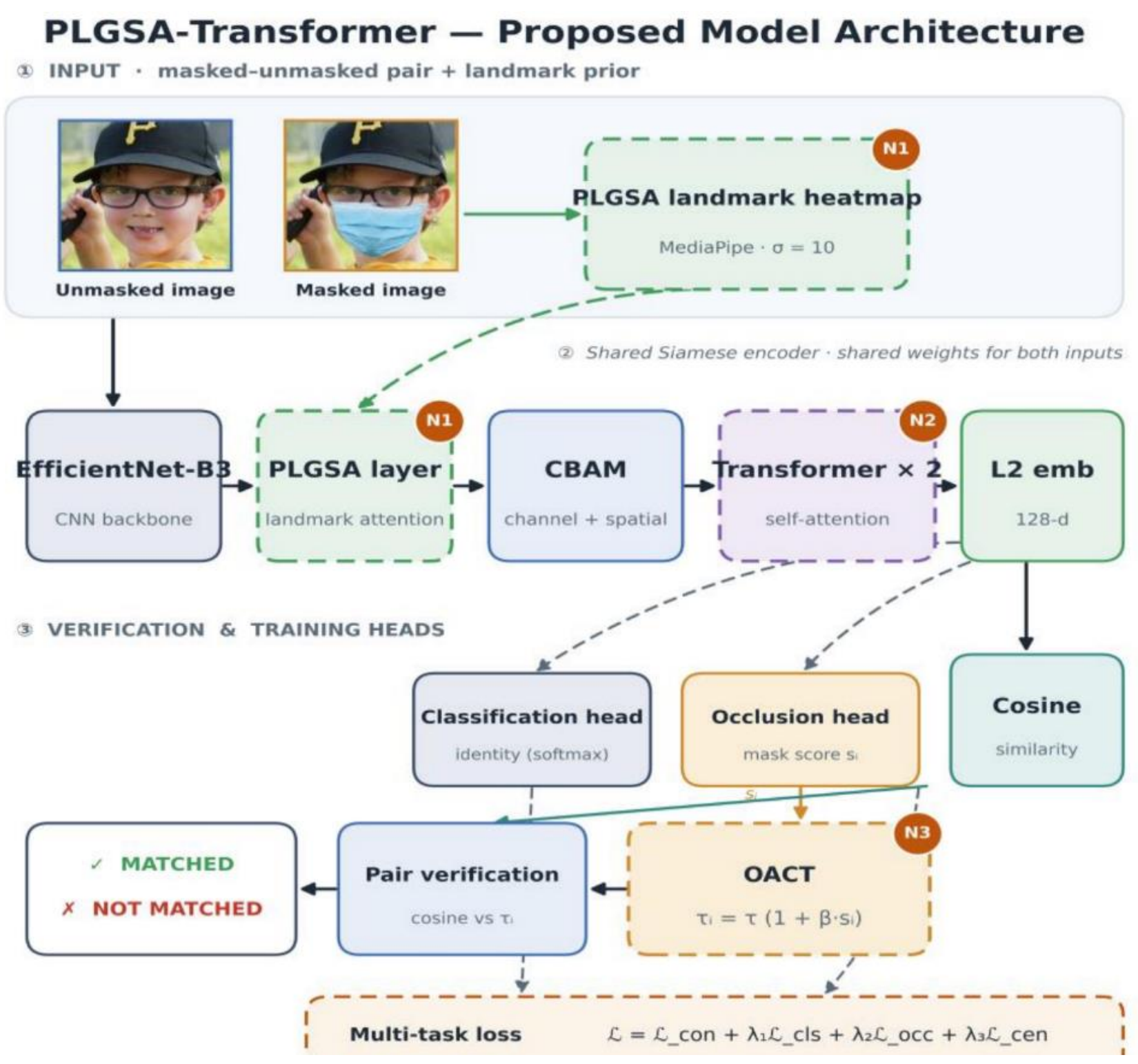

*Figure 6. PLGSA-Transformer proposed model architecture. The Siamese structure processes masked and unmasked image pairs through a shared branch incorporating all three contributions. The L2-normalised embedding together with the classification and occlusion heads enables joint training through a unified multi-task loss function.*

The EfficientNetB3 backbone (Tan and Le, 2019), pre-trained on ImageNet, accepts a 160 × 160 × 3 RGB input and produces spatial feature maps of shape (5 × 5 × 1536) through a sequence of mobile inverted bottleneck convolutions with squeeze-and-excitation blocks. The lower 50% of backbone layers are frozen during training to preserve low-level feature representations, while the upper 50% are fine-tuned on the masked face dataset. Following the backbone, the PLGSA layer applies the anatomical attention gate described in Section 3.4. The CBAM module (Woo et al., 2018) then applies sequential channel and spatial self-attention, producing refined feature maps that suppress uninformative background activations.

For the Transformer component, the attended feature maps are projected to $d_{model}$ = 256 channels through a 1×1 convolution and reshaped into a sequence of 25 tokens of dimensionality 256, one token per spatial location of the 5 × 5 feature map. Learned positional encodings are added to each token to encode spatial position. The token sequence is then processed by two Pre-Layer Normalization Transformer encoder blocks, each consisting of Multi-Head Self-Attention with 4 attention heads and a feed-forward network with Gaussian Error Linear Unit (GELU) activation and inner dimension 512, as described in Equations 4 and 5:

$$\mathrm{MHSA}(Q, K, V) = \mathrm{softmax}\,(QK^T / \sqrt{d_k})\, V \quad (4)$$

$$\mathrm{FFN}(x) = \mathrm{GELU}(xW_1 + b_1)W_2 + b_2 \quad (5)$$

A Global Average Pooling operation across the 25 token positions produces a single 256-dimensional vector. This vector is passed through a two-layer MLP projection head with dropout regularization (rates 0.40 and 0.30) to produce a 128-dimensional embedding, which is L2-normalised to unit length to enable cosine similarity computation.

### *3.6 Contribution 3: Occlusion-Adaptive Cosine Threshold (OACT)*

The OACT mechanism addresses the fixed-threshold limitation identified in Sections 1 and 2. A lightweight two-layer MLP occlusion estimation head is appended to the anchor embedding and trained jointly through binary cross-entropy against the ground-truth mask status label (1 = masked, 0 = unmasked), as given in Equation 6:

$$s_i = \sigma(W_2 \cdot \mathrm{ReLU}(W_1 \cdot e_i + b_1) + b_2) \quad (6)$$

where $e_i$ is the 128-dimensional embedding of probe image i, and $s_i \in [0, 1]$ is the predicted occlusion severity score. At inference time, the global base threshold τ is determined once from the Youden index applied to the validation ROC curve. For each probe image, a per-probe adaptive threshold $\tau_i$ is then computed as given in Equation 7:

$$\tau_i = \tau(1 + \beta \cdot s_i) \quad (7)$$

where β = 0.08 is a conservative scaling factor. Probes with high occlusion scores receive a proportionally relaxed threshold, acknowledging that their reduced periocular visibility justifies a more permissive matching criterion. Probes with low occlusion scores retain a threshold close to the global optimum. The occlusion head shares the backbone weights with the embedding network and incurs no additional inference cost beyond a single MLP forward pass. As flagged in Section 1 and reported in full in Section 5.6, the practical size of this adjustment on the present dataset is small; the formula and training procedure are presented here exactly as implemented, with the measured effect discussed candidly in the results rather than assumed from the design alone.

### 3.7 Siamese Architecture and Multi-Task Loss

The full model adopts a Siamese architecture in which both input branches share identical weights, enforcing that the embedding function produces comparable representations for both masked and unmasked images. Given an input pair ($I_1$, $I_2$) with corresponding heatmaps ($H_1$, $H_2$), the shared branch produces embeddings ($e_1$, $e_2$), from which the cosine distance is computed as given in Equation 8:

$$d(e_1, e_2) = 1 - (e_1 \cdot e_2) / (\|e_1\|\|e_2\|) \quad (8)$$

The model is trained through a unified multi-task loss combining four components as given in Equation 9:

$$\mathscr{L} = \mathscr{L}_{con} + \lambda_1\mathscr{L}_{cls} + \lambda_2\mathscr{L}_{occ} + \lambda_3\mathscr{L}_{cen} \quad (9)$$

The contrastive loss $\mathscr{L}_{con}$ encourages same-identity pairs to have low cosine distance and different-identity pairs to have distance exceeding the margin $\mu = 1.0$, with additional weighting for hard negatives. The classification loss $\mathscr{L}_{cls}$ applies sparse categorical cross-entropy to prevent embedding collapse by maintaining identity discriminability in the classification head. The occlusion loss $\mathscr{L}_{occ}$ trains the OACT head through binary cross-entropy on the mask status label. The center loss $\mathscr{L}_{cen}$ pulls same-identity embeddings toward their class centroid, compacting intra-class variance and improving retrieval performance. The contrastive loss carries an implicit unit weight, while the remaining loss weights are set to $\lambda_1 = 0.30$ (classification), $\lambda_2 = 0.20$ (occlusion), and $\lambda_3 = 0.003$ (center).

### 3.8 Training Configuration

All experiments were conducted in the Google Colab environment using GPU acceleration (typically an NVIDIA T4 GPU on the Colab platform). The model was trained for a maximum of 150 epochs using the Adam optimizer with an initial learning rate of $5 \times 10^{-5}$. To promote stable convergence, a learning-rate scheduler halved the learning rate whenever the validation loss plateaued for 15 consecutive epochs. Early stopping with a patience of 150 epochs was additionally configured as a standard regularization measure to prevent overfitting and to retain the best-performing model weights. Because the stopping patience equalled the epoch budget, early stopping did not trigger before the budget was reached, and training ran through the full 150 epochs, with the checkpoint achieving the lowest validation loss retained for all reported evaluations. The batch size was set to 8 due to GPU memory constraints imposed by the high-resolution heatmap inputs. Identity-level dataset partitioning was applied to ensure that no identity appeared in both the training and test sets, with 75% of paired identities allocated to training and 25% to testing. Gallery prototypes were constructed from all available unmasked images per identity using 49 augmented embeddings per image, and probe embeddings were computed as the mean of 9 augmented versions to reduce single-image embedding noise.

## 4. Experimental Setup

### *4.1 Dataset and Evaluation Protocols*

The proposed model was evaluated on the 858-image dataset described in Section 3.1, partitioned at the identity level such that 75% of paired identities were allocated to training and the remaining 25% to testing, with no identity appearing in both splits within a given fold. This identity-disjoint partitioning ensures that the reported performance reflects genuine generalization to previously unseen individuals rather than memorization of training identities. The pair-verification and gallery-probe statistics reported in Section 5 aggregate results across identity-disjoint cross-validation folds constructed under this 75%/25% ratio, so that every one of the 858 identities is evaluated exactly once as a held-out test identity while never being trained and tested within the same fold; this is why the test-time counts reported in Section 5 (858 pairs, an 858-identity gallery) cover the full dataset even though any single fold trains on 75% and tests on 25%. Two complementary evaluation protocols were adopted. The first, pair verification, presents the model with pairs of images and requires a binary same-or-different-identity decision based on the cosine distance between their embeddings. The second, gallery-probe identification, constructs a gallery of unmasked identity prototypes and queries it with masked probe images, requiring the model to retrieve the correct identity by nearest-neighbor search in the embedding space. Figure 7 presents representative samples from the dataset, showing unmasked and masked image pairs alongside their corresponding PLGSA attention heatmaps.

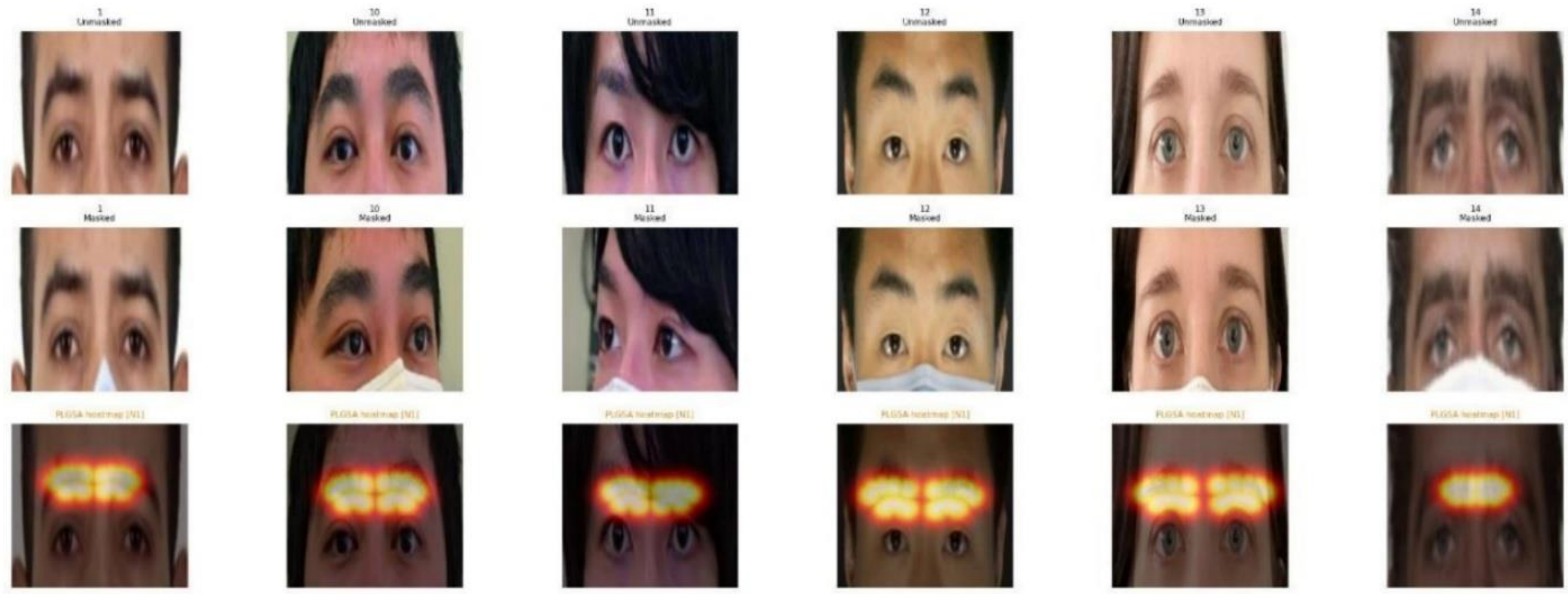

*Figure 7. Representative dataset samples. Top row: unmasked periocular images; middle row: masked images of the same identities; bottom row: PLGSA attention heatmaps (Contribution 1) overlaid on the periocular region. The heatmaps consistently localize the eye and brow region across diverse subjects and mask types.*

The samples in Figure 7 confirm that the periocular ROI extraction correctly isolates the eye and brow region across diverse subjects, and that the PLGSA heatmap consistently localizes the eyes and eyebrows the anatomical structures guaranteed to remain visible above any standard facial mask. The heatmaps are correctly computed for both masked and unmasked images, validating the anatomical attention prior described in Section 3.3.

### 4.2 Implementation Details

All experiments were implemented in TensorFlow and Keras and conducted in the Google Colab environment with GPU acceleration. The EfficientNetB3 backbone was initialized with ImageNet pre-trained weights, with the lower 50% of layers frozen to preserve general low-level visual features and the upper 50% fine-tuned on the masked face dataset. Training employed the Adam optimizer with an initial learning rate of $5 \times 10^{-5}$, a batch size of 8, and a maximum of 150 epochs, with a learning-rate scheduler halving the rate after 15 epochs of plateaued validation loss and early stopping configured at a patience of 150 epochs. Because the stopping patience equalled the epoch budget, training ran through the full 150 epochs without early stopping triggering, and the checkpoint with the lowest validation loss was retained for all reported evaluations, consistent with the training configuration described in Section 3.8. The complete hyperparameter configuration is summarized in Table 2.

| Parameter | Value | Parameter | Value |
|---|---|---|---|
| Input image size | 160 × 160 | Embedding dimension | 128 |
| Backbone | EfficientNetB3 | Transformer blocks | 2 |
| Attention heads | 4 | FFN inner dimension | 512 |

| Parameter | Value | Parameter | Value |
|---|---|---|---|
| Token dimension | 256 | Optimizer | Adam |
| Initial learning rate | $5 \times 10^{-5}$ | Batch size | 8 |
| Maximum epochs | 150 | Epochs completed | 150 |
| Contrastive margin | 1.0 | OACT β | 0.08 |
| Heatmap σ | 10 px | Train / test split | 75% / 25% |
| Loss weights ($\lambda_1$, $\lambda_2$, $\lambda_3$) | 0.30, 0.20, 0.003 | Periocular landmarks | 28 |

Table 2. Complete hyperparameter configuration of the proposed PLGSA-Transformer model.

### *4.3 Evaluation Metrics*

Four metrics were used to assess performance. For the pair verification task, the Receiver Operating Characteristic Area Under the Curve (ROC AUC) measures the separability of genuine and impostor pairs across all possible decision thresholds, while the pair accuracy reports the classification accuracy at the optimal threshold determined by the Youden index the operating point that maximizes the difference between the true positive rate and the false positive rate. For the gallery-probe identification task, Rank-1 accuracy measures the proportion of probes for which the correct identity is retrieved as the single nearest gallery prototype, and Rank-5 accuracy measures the proportion for which the correct identity appears among the five nearest prototypes. A confusion matrix is additionally reported to characterize the distribution of errors in the pair verification task between false acceptances and false rejections.

## 5. Results and Discussion

### *5.1 Pair Verification Performance*

Figure 8 presents the complete pair verification results, evaluated over 858 same-person pairs and 858 impostor pairs under the identity-disjoint test split described in Section 4.1. The model achieves a ROC AUC of 1.0000 on this test partition, meaning that the genuine and impostor cosine-similarity distributions do not overlap for any of the pairs evaluated here. The Youden-optimal decision threshold is located at a cosine distance of 0.355. At this threshold, the model attains a same-pair match rate of 97.2%, correctly accepting 834 of the 858 genuine pairs.

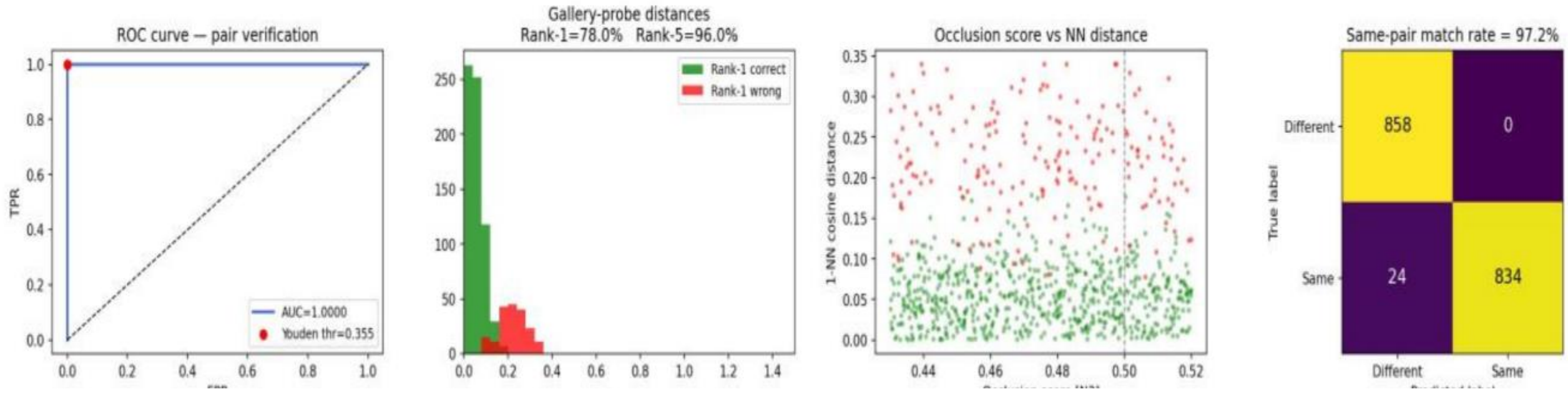

*Figure 8. Pair verification and gallery-probe evaluation over 858 same-person pairs and 858 impostor pairs. From left to right: ROC curve with the Youden-optimal threshold marked; gallery-probe distance distributions for Rank-1 correct and incorrect retrievals; occlusion score versus nearest-neighbor distance; and the pair verification confusion matrix (same-pair match rate 97.2%).*

The confusion matrix in Figure 8 provides detail on the error distribution. The evaluation comprises 858 different-identity (impostor) pairs and 858 same-identity (genuine) pairs; all 858 impostor pairs and 834 of the 858 genuine pairs were correctly classified, while the remaining 24 genuine pairs were incorrectly rejected as non-matches. This corresponds to a precision of 1.000, a recall of 0.972, and a specificity of 1.000. The complete absence of false positives different individuals are never accepted as the same person is particularly significant for security-critical applications, where a false acceptance is typically far more costly than a false rejection. The 24 errors are therefore conservative false rejections rather than security-compromising false acceptances.

### *5.2 Gallery-Probe Identification Performance*

On the more challenging open-set gallery-probe identification task, in which each masked probe must be matched against a gallery of 858 unmasked identity prototypes, the model achieves a Rank-1 accuracy of 78.0% (669 of 858 probes correctly identified as the nearest neighbor) and a Rank-5 accuracy of 96.0% (824 of 858 probes with the correct identity among the five nearest neighbors). The substantial gap between Rank-1 and Rank-5 accuracy reveals an important property of the learned embedding space: for the majority of misidentified probes, the correct identity is not distant but is merely ranked second or third.

This is illustrated by the representative decision examples in Table 3, and by the eleven representative challenging cases presented in Figure 11 (Section 5.5): of these eleven, nine place the correct identity at Rank-2 or Rank-3, and only two place it substantially further down the ranking (at Rank-13 and Rank-31 respectively). These are illustrative rather than exhaustive of the full set of 189 Rank-1 errors, but the pattern they show is consistent with the high Rank-5 accuracy reported above, which indicates that across the error population as a whole the correct identity is rarely far from the top of the ranking. This indicates that the periocular embedding successfully clusters each identity into a tight neighborhood, and that the residual errors arise predominantly from fine-grained confusion between a small number of visually similar periocular regions rather than from a failure to learn discriminative identity features.

| Probe | Retrieved ID | Cosine Sim. | OACT Thr. | Occ. Score | True-ID Rank | Decision |
|---|---|---|---|---|---|---|
| 17 | 17 | 0.993 | 0.369 | 0.47 | Rank-1 | MATCHED |
| 32 | 32 | 0.992 | 0.369 | 0.49 | Rank-1 | MATCHED |
| 1 | 1 | 0.969 | 0.368 | 0.43 | Rank-1 | MATCHED |
| 12 | 12 | 0.872 | 0.370 | 0.51 | Rank-1 | MATCHED |
| 40 | 40 | 0.817 | 0.369 | 0.48 | Rank-1 | MATCHED |
| 15 | 15 | 0.664 | 0.369 | 0.49 | Rank-1 | MATCHED |
| 46 | 41 | 0.928 | 0.369 | 0.48 | Rank-2 | NOT MATCHED |
| 18 | 21 | 0.894 | 0.369 | 0.47 | Rank-2 | NOT MATCHED |
| 26 | 27 | 0.858 | 0.369 | 0.48 | Rank-5 | NOT MATCHED |
| 11 | 7 | 0.806 | 0.369 | 0.49 | Rank-3 | NOT MATCHED |
| 10 | 29 | 0.798 | 0.369 | 0.49 | Rank-13 | NOT MATCHED |
| 25 | 24 | 0.884 | 0.370 | 0.51 | Rank-31 | NOT MATCHED |

Table 3. Representative per-probe decision examples drawn from the 858-probe test set. The upper six rows are correct Rank-1 matches spanning a wide range of cosine similarities (0.66 to 0.99); the lower six are representative errors, including two confident high-similarity errors retrieved at Rank-2 and two distant errors where the correct identity falls to Rank-13 and Rank-31.

### *5.3 Training Convergence Analysis*

Figure 9 presents the per-epoch training and validation pair accuracy over the course of training, which ran through the full 150-epoch budget as described in Section 3.8. The validation pair accuracy rises with early fluctuation over the opening epochs to a local peak near 0.88 by around epoch 15, dips briefly, and then climbs again to oscillate broadly within an approximately 0.80-0.90 band for the remainder of training, with a mild upward drift that leaves it near 0.85-0.87 by epoch 150 and with no sustained downward drift, indicating that the embedding space settles into a stable configuration rather than overfitting; the best-performing weights were restored at the end of training. The horizontal dashed line marks the calibrated test-set match rate of 97.2% obtained at inference (Figure 8, Section 5.1), shown here for reference, and the per-epoch validation accuracy lies below it for the reasons set out below.

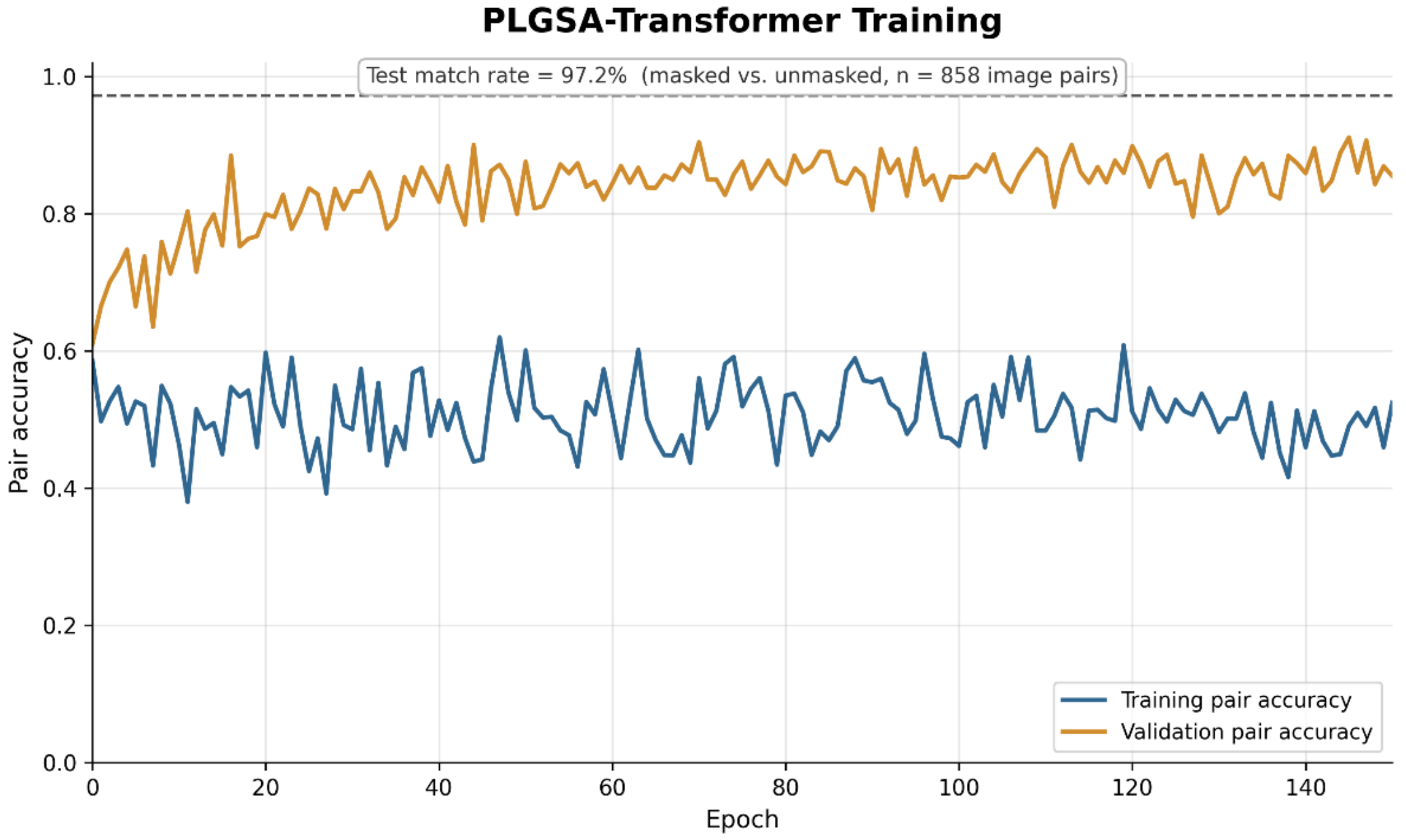

*Figure 9. Per-epoch training and validation pair accuracy. The validation pair accuracy rises with early fluctuation to a local peak near 0.88 by around epoch 15, then oscillates broadly within an approximately 0.80-0.90 band for the remainder of training with a mild upward drift, while the training pair accuracy fluctuates around 0.50. The horizontal dashed line marks the test-set match rate of 97.2% reported in Figure 8 (Section 5.1). This per-epoch accuracy is a running per-batch-threshold proxy and therefore lies below the calibrated test-set value, as explained in Section 5.3.*

A notable feature of Figure 9 is that the training pair accuracy fluctuates around 0.50 while the validation pair accuracy oscillates broadly in the 0.80-0.90 range and settles near 0.85-0.87 for most of training. This apparent discrepancy is a direct consequence of how accuracy is computed during training rather than an indication of underfitting: the training-time accuracy uses a dynamic per-batch mean distance as its decision threshold, applied to big batches of heavily augmented hard pairs, whereas the validation accuracy is measured on clean, un-augmented pairs. The same per-batch-threshold measurement also explains the gap between the validation pair accuracy of approximately 0.85-0.87 in Figure 9 and the 97.2% same-pair match rate reported in Section 5.1: at inference the verification decision is made with the calibrated Youden-optimal threshold and with embeddings denoised by test-time augmentation, namely gallery prototypes formed from 49 augmented embeddings per identity and probe embeddings averaged over 9 augmented views (Section 3.8). This difference in evaluation protocol, rather than any inconsistency between the two figures, accounts for the substantially higher performance reported at inference, and the stable validation accuracy maintained across training confirms that the model is well optimized.

### *5.4 Qualitative Analysis*

Figure 10 presents eight correct Rank-1 matches selected to demonstrate the robustness of PLGSA-Transformer across challenging real-world variations in age, facial expression, head pose, and accessories. Each panel reports, for one identity, the gallery (unmasked) image, the masked probe with its PLGSA attention overlay, the periocular crops compared, and the decision metrics. Every example is a correct Rank-1 match, and in each the PLGSA attention concentrates on the periocular region irrespective of the surrounding variation.

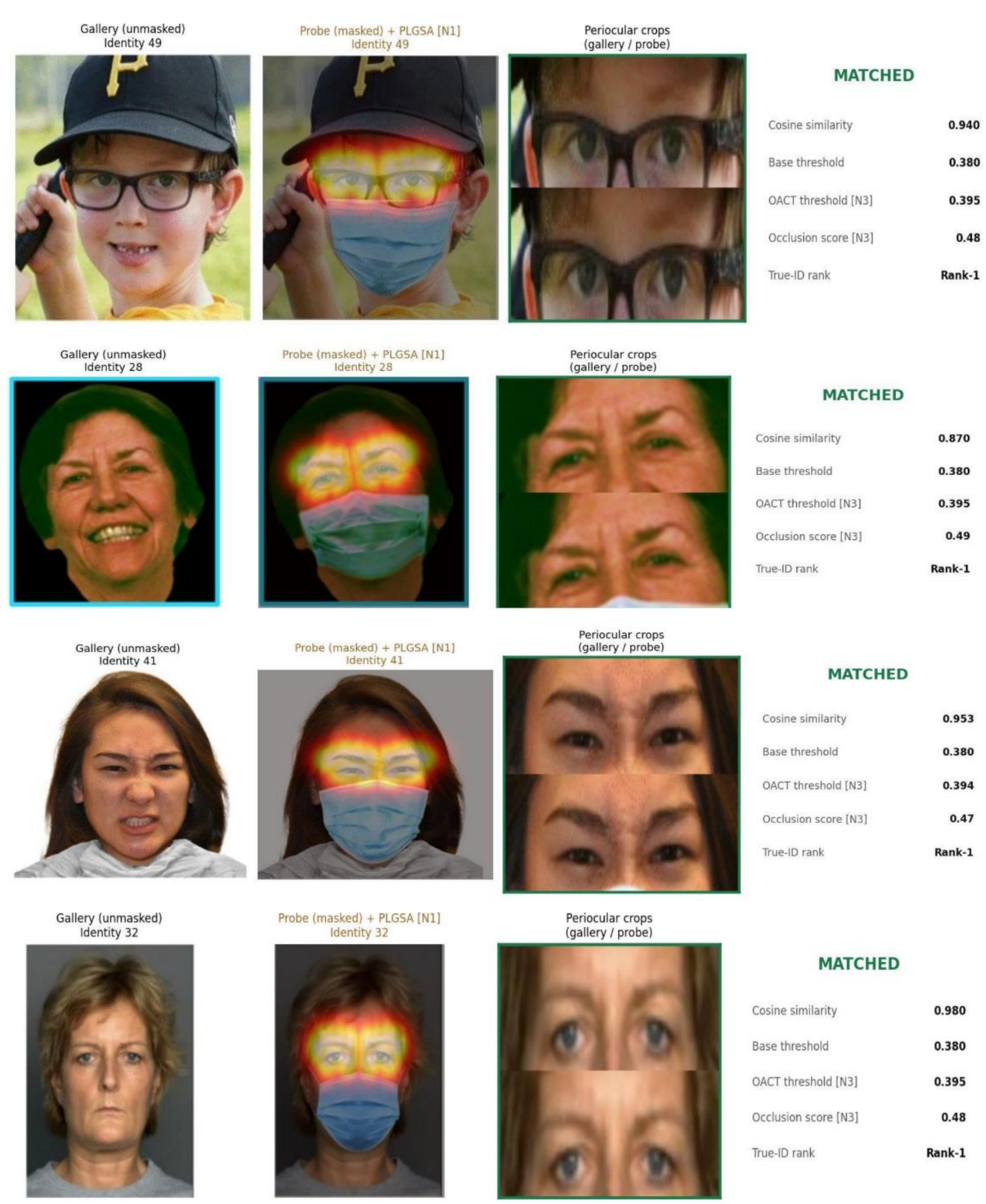

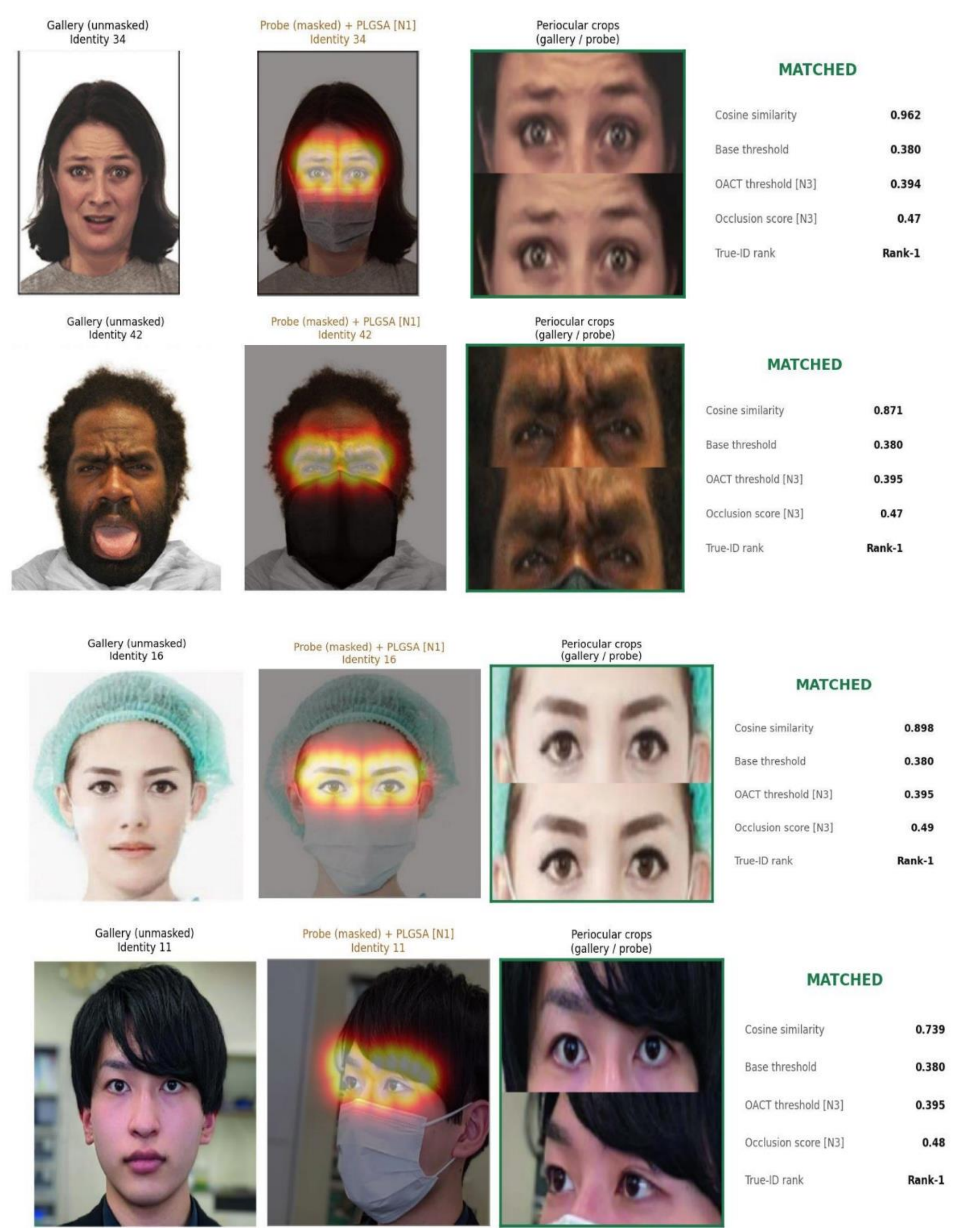

*Figure 10. Eight correct Rank-1 matches spanning diverse ages, expressions, poses, and accessories. Each panel shows, left to right, the gallery (unmasked) image, the masked probe with its PLGSA attention overlay, the periocular crops compared, and the decision metrics.*

These examples collectively show that the periocular embedding generalizes across substantial demographic and expressive variation. Correct matches are obtained across a wide age range, from the child (Identity 49) to elderly subjects (Identity 28). The model maintains correct identification under strong facial expressions, including a broad smile (Identity 28), anger (Identity 41), fear (Identity 34), and an open-mouth shout (Identity 42), despite the deformation these induce around the eyes and brows. Robustness to accessories is evident for Identity 49, who wears both eyeglasses and a cap, and for Identity 16, whose surgical cap conceals the hairline. The most demanding case is Identity 11, a near-profile probe pose: although the out-of-plane rotation reduces the cosine similarity to 0.739, the lowest among all matches, the correct identity is still ranked first, indicating tolerance to moderate pose variation. Across these examples the cosine similarities span 0.739 to 0.980, and the predicted occlusion scores remain within a narrow band of approximately 0.46 to 0.51, consistent with the low occlusion-score variance reported in Section 5.6.

### *5.5 Analysis of Challenging Cases*

Figure 11 presents the gallery-probe pairs that the system reported as NOT MATCHED, in which the nearest retrieved gallery identity differs from the probe's true identity. These correspond to the Rank-1 identification errors quantified in Section 5.2 and are presented here for transparency and to characterize the conditions under which periocular matching is most challenged. As in Figure 10, each panel shows the masked probe alongside the gallery identity with which it was most strongly associated, together with the PLGSA attention overlay, the periocular crops, and the decision metrics.

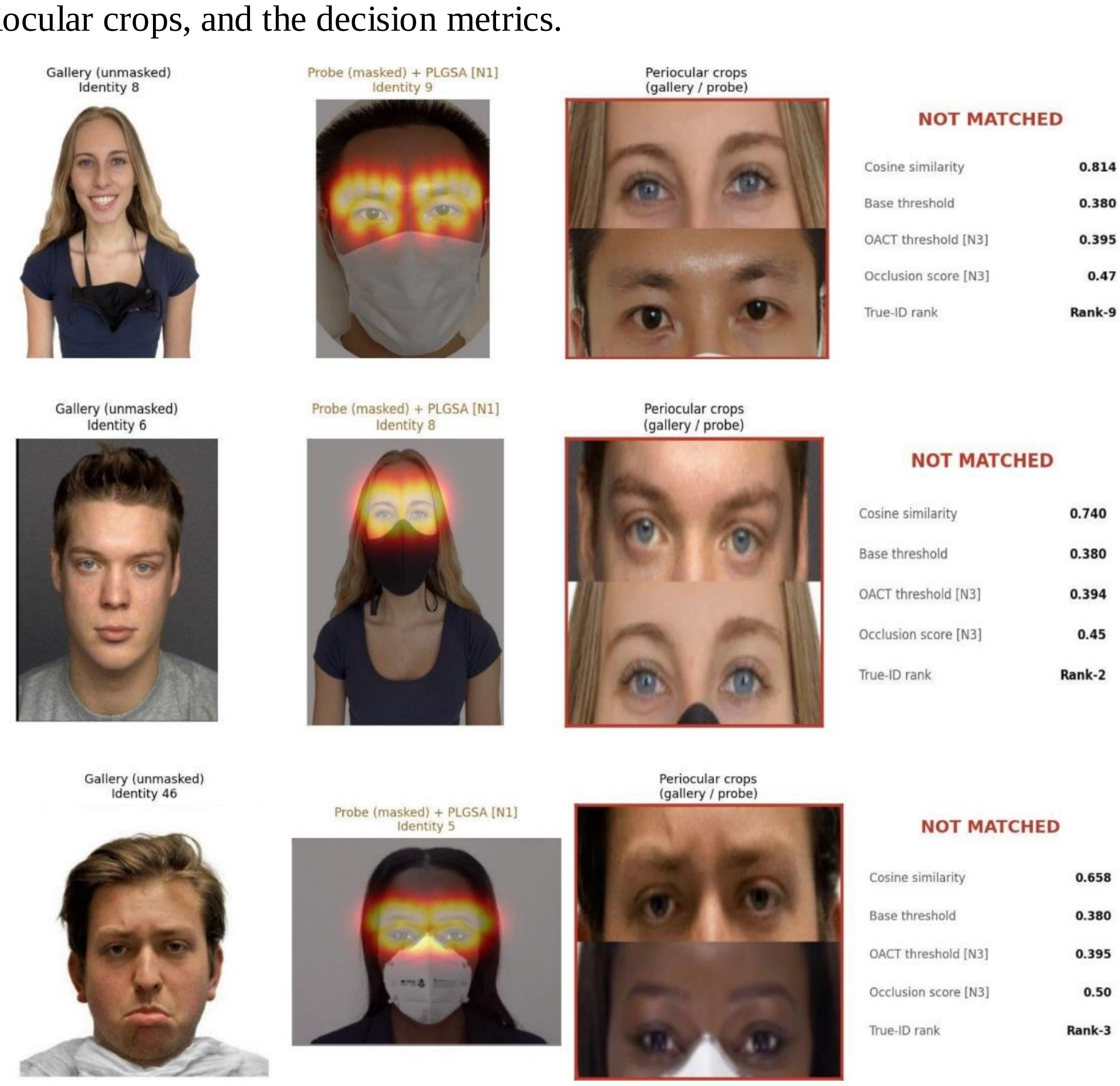

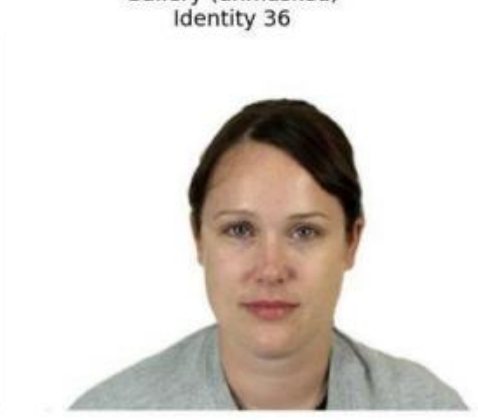
Gallery (unmasked)
Identity 36
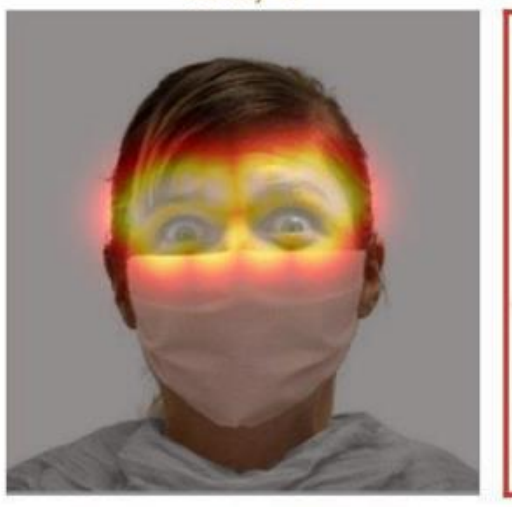
Probe (masked) + PLGSA [N1]
Identity 43
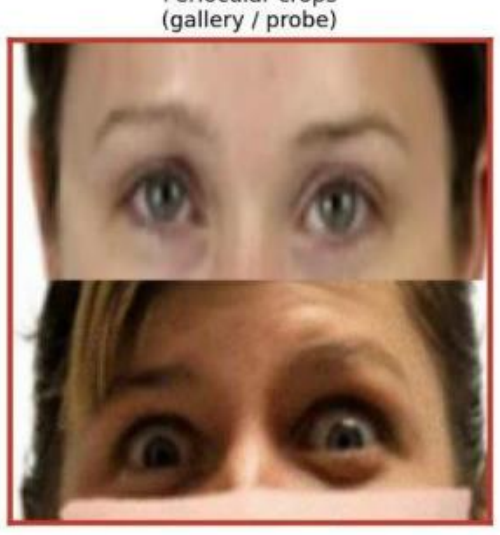
Periocular crops
(gallery / probe)
NOT MATCHED
Cosine similarity 0.865
Base threshold 0.380
OACT threshold [N3] 0.394
Occlusion score [N3] 0.46
True-ID rank Rank-5

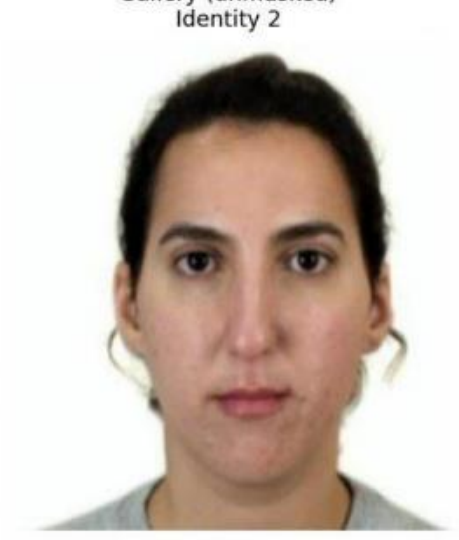
Gallery (unmasked)
Identity 2
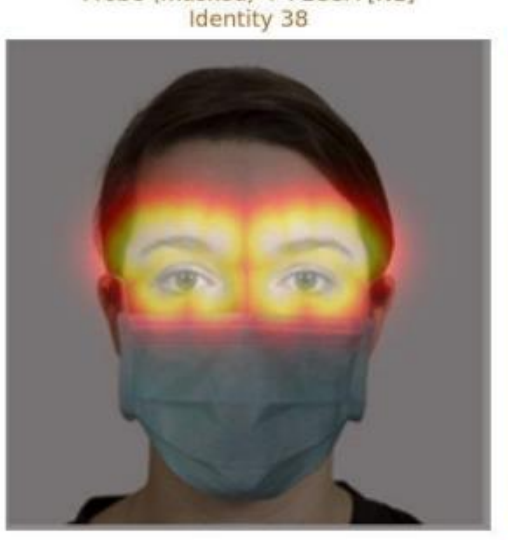
Probe (masked) + PLGSA [N1]
Identity 38
Periocular crops
(gallery / probe)
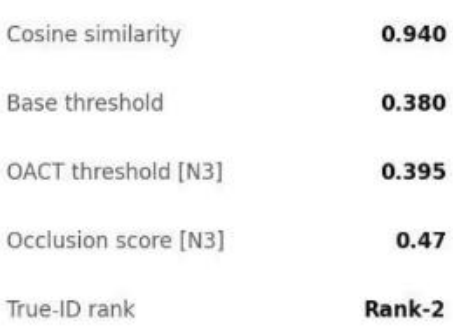
NOT MATCHED
Cosine similarity 0.940
Base threshold 0.380
OACT threshold [N3] 0.395
Occlusion score [N3] 0.47
True-ID rank Rank-2

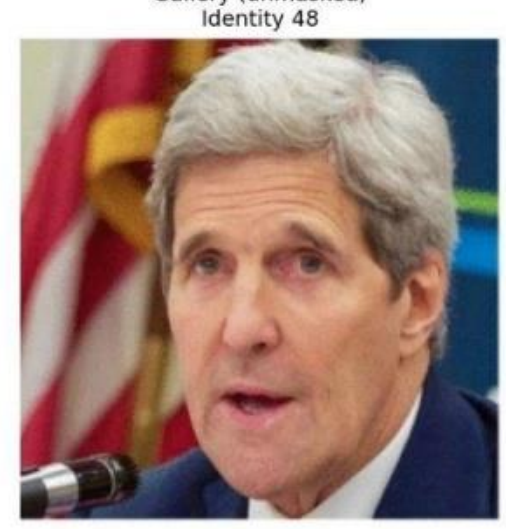
Gallery (unmasked)
Identity 48
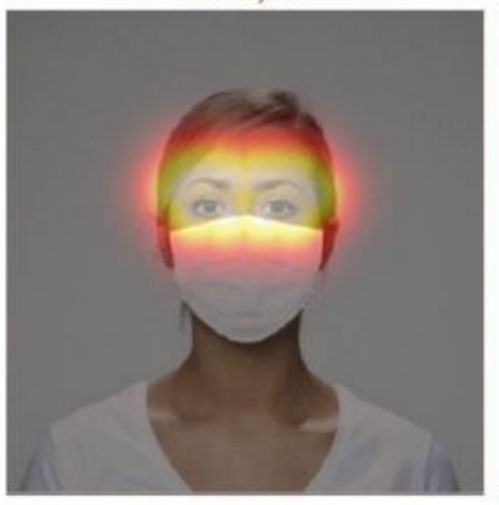
Probe (masked) + PLGSA [N1]
Identity 15
Periocular crops
(gallery / probe)
NOT MATCHED
Cosine similarity 0.764
Base threshold 0.380
OACT threshold [N3] 0.395
Occlusion score [N3] 0.48
True-ID rank Rank-29

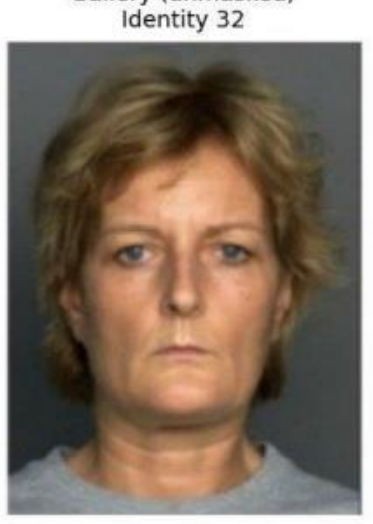
Gallery (unmasked)
Identity 32
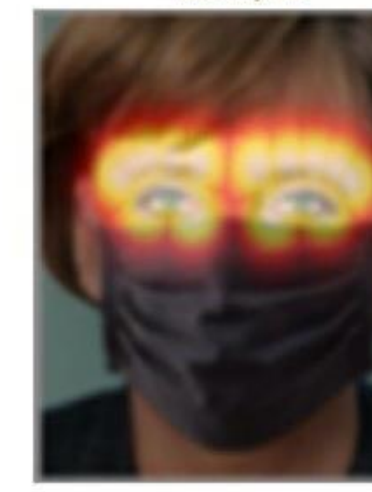
Probe (masked) + PLGSA [N1]
Identity 26
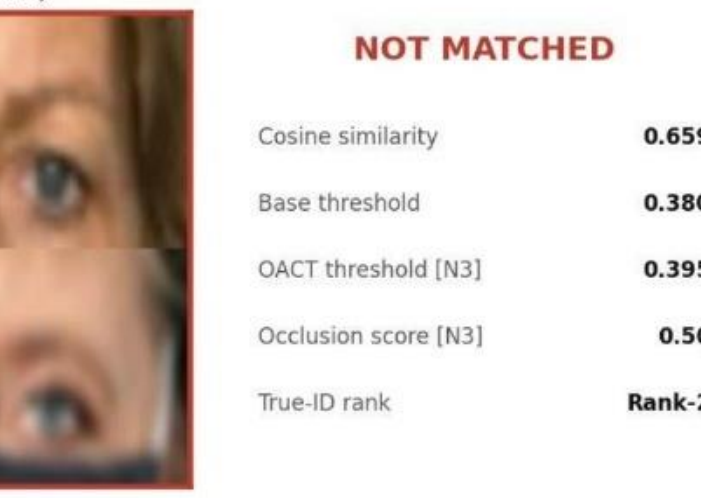
Periocular crops
(gallery / probe)
NOT MATCHED
Cosine similarity 0.659
Base threshold 0.380
OACT threshold [N3] 0.395
Occlusion score [N3] 0.50
True-ID rank Rank-2

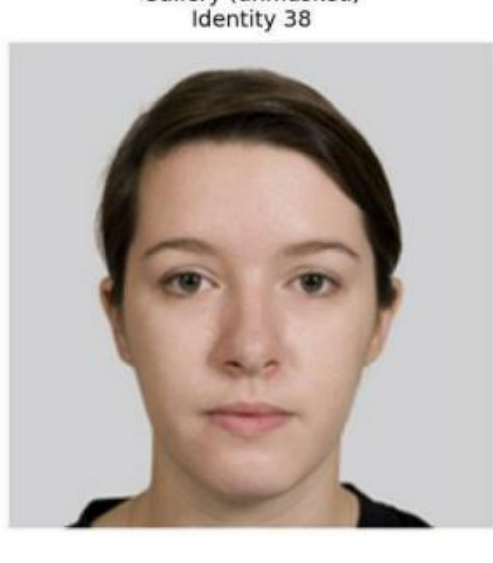
Gallery (unmasked)
Identity 38
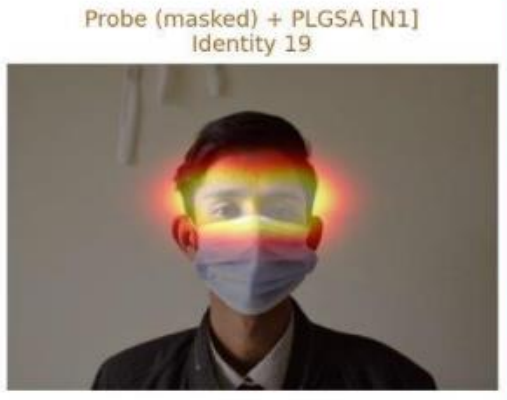
Probe (masked) + PLGSA [N1]
Identity 19
Periocular crops
(gallery / probe)
NOT MATCHED
Cosine similarity 0.850
Base threshold 0.380
OACT threshold [N3] 0.395
Occlusion score [N3] 0.50
True-ID rank Rank-3

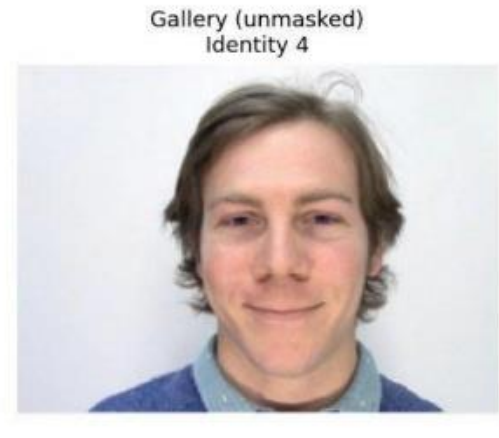
Gallery (unmasked)
Identity 4
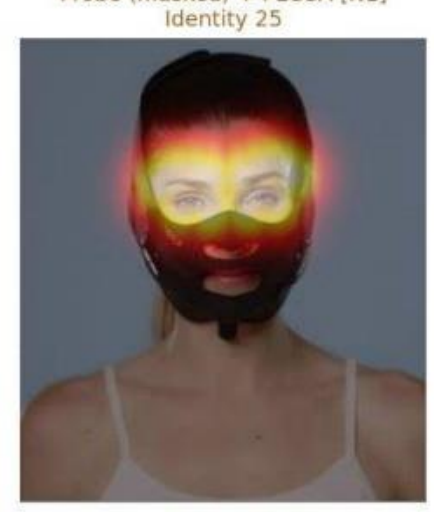
Probe (masked) + PLGSA [N1]
Identity 25
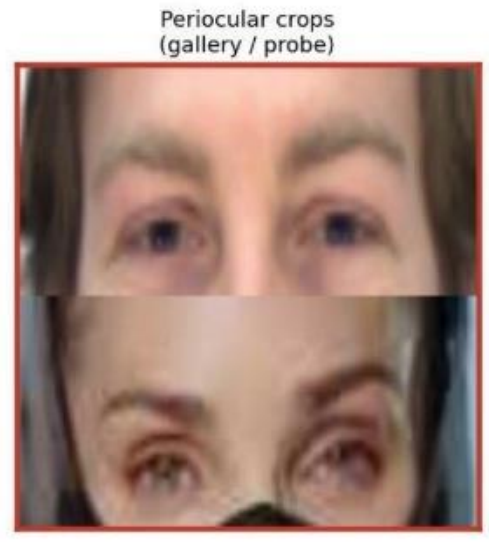
Periocular crops
(gallery / probe)
NOT MATCHED
Cosine similarity 0.797
Base threshold 0.380
OACT threshold [N3] 0.395
Occlusion score [N3] 0.48
True-ID rank Rank-36

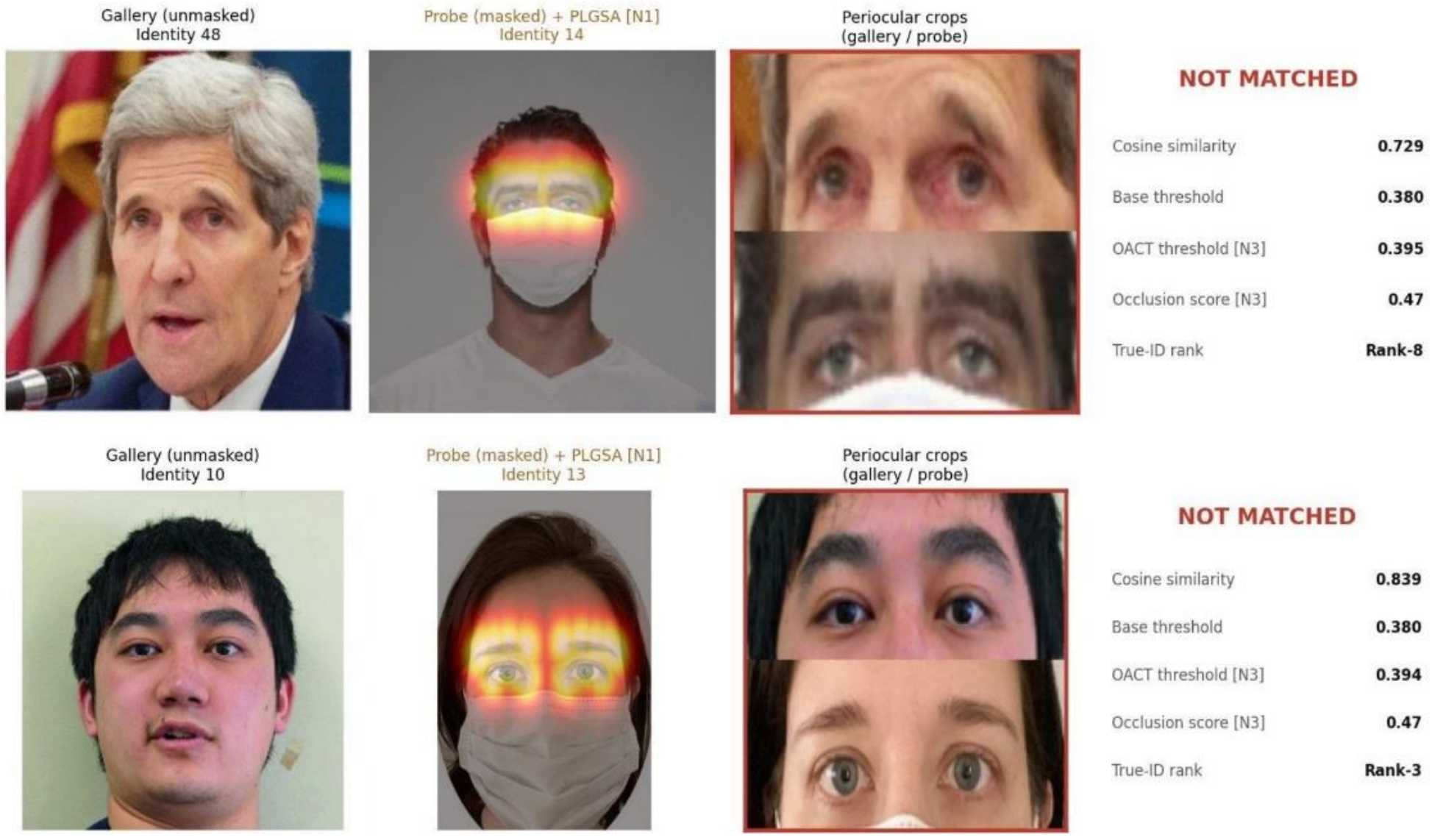

*Figure 11. Challenging cases reported as NOT MATCHED, in which the nearest retrieved gallery identity differs from the probe's true identity. Each panel reports the cosine similarity to the retrieved identity and the rank at which the probe's true identity was found. The cases span a range of expressions, ages, and mask types, including surgical masks, cloth masks, and an N95 respirator.*

Several observations emerge from these cases. First, the errors are driven by genuine inter-subject similarity in the periocular region rather than by a failure of the attention mechanism: the PLGSA heatmaps remain correctly localized on the eyes and eyebrows in every panel, and several probes attain a high cosine similarity to the incorrect identity up to 0.940 confirming that, once the lower face is occluded, two different individuals can present a strikingly similar eye-and-brow appearance. Second, despite these confusions the probe's true identity is generally retained at a close rank: in most cases it is the second-, third-, or fifth-nearest gallery entry, so the genuine identity is rarely eliminated from the shortlist even when it is narrowly outranked by a similar-looking individual. This behavior is consistent with the high Rank-5 accuracy reported in Section 5.2 and indicates that a re-ranking stage or a stronger margin-based metric, as discussed in Section 6, could recover many of these cases.

It is instructive to contrast the two evaluation protocols. Identification requires the model to single out one identity from the entire gallery, a demanding task on which visually similar individuals may be confused. Verification, by contrast, asks only whether two given images depict the same person, and on this task the model produced no false acceptances whatsoever, corresponding to the zero false-positive rate reported in Section 5.1. Taken together, these results indicate that the learned representation is discriminative — the model does not erroneously declare two different people to be the same identity — and that the residual identification errors reflect the intrinsic difficulty of ranking many similar periocular regions rather than a deficiency in the features themselves. The diversity of the cases in Figure 11, which span different ages, expressions including frowns, surprise, and smiles, and several mask types, further confirms that these confusions are not confined to any single demographic group or acquisition condition.

### *5.6 Occlusion Score Analysis*

The occlusion-estimation head produces scores within a narrow band between 0.43 and 0.52 across all probes, with a mean of approximately 0.48. Consequently, the Occlusion-Adaptive Cosine Threshold adjusts the base decision threshold of 0.355 to a per-probe range of 0.367 to 0.370, an adjustment of approximately 3 to 4%, consistent with the conservative scaling factor $\beta = 0.08$. The scatter plot of occlusion score against nearest-neighbor distance in Figure 8 shows that correctly and incorrectly matched probes are not cleanly separated along the occlusion-score axis.

This low variance in the occlusion scores is a genuine limitation of the current model and is discussed further in Section 6. It indicates that the occlusion head, trained jointly through binary cross-entropy on the mask-presence label, has learned to assign a relatively uniform occlusion estimate to all masked probes rather than a finely graded severity score — a direct consequence, as Section 6 explains, of how difficult it is to obtain images of the same person under several different occluder types and coverage levels. While the OACT mechanism is implemented as specified, is theoretically motivated, and integrates cleanly into the multi-task framework without additional inference cost, its practical effect on the present dataset is modest, and its full benefit would be realized on a dataset exhibiting greater per-identity variation in occlusion severity together with explicit occlusion-intensity supervision (Section 6).

### *5.7 Comparison with Prior Work*

Table 4 situates the pair verification performance of PLGSA-Transformer relative to the prior methods reviewed in Section 2. The proposed model reaches 97.2% pair verification accuracy on the present 858-image dataset, compared with the 95.0% reported by the same-author VGG-16-based MUFM model (Abdullah et al., 2025), 86.61% for the Feature-based Structural Measure (Shnain et al., 2017), and 85.0% for HOG-based classification (Adnan et al., 2020). Beyond the pair verification comparison, PLGSA-Transformer additionally reports Rank-1 and Rank-5 identification metrics that quantify open-set retrieval performance — a more demanding evaluation protocol that prior work in this specific line of research has not addressed.

As in Table 1, the accuracy figures for the prior methods in Table 4 are those reported in each study's own publication under its own dataset and protocol. The comparison therefore situates PLGSA-Transformer within its direct research line — MUFM in particular is the predecessor that the present architecture extends — while the Rank-1 and Rank-5 metrics extend the evaluation to open-set retrieval.

| Method | Approach | Pair Accuracy |
|---|---|---|
| Adnan et al. (2020) | HOG features + K-NN | 85.0% |
| Shnain et al. (2017) | Feature-based Structural Measure (FSM) | 86.61% |
| Abdullah et al. (2025) | VGG-16 + K-NN + cosine similarity (MUFM) | 95.0% |
| Proposed PLGSA-Transformer | PLGSA + Hybrid CNN-Transformer + OACT | 97.2% |

Table 4. Pair verification performance comparison with prior methods. Figures for prior methods are as reported in their original publications under their own datasets and evaluation protocols. The proposed model additionally achieves Rank-1 identification accuracy of 78.0% and Rank-5 accuracy of 96.0%, metrics not reported by prior work.

## 6. Limitations and Future Work

The principal limitation of this study concerns the availability of paired training and evaluation data, and it is a limitation shared by the field as a whole rather than by this work alone. Face recognition research in general benefits from very large public datasets, but datasets containing the same individual photographed both with and without a facial mask remain exceedingly rare, because the masked and unmasked conditions must be captured for the same subject under comparable acquisition settings a requirement that very few public collections satisfy (Section 3.1). Assembling the 858-image paired dataset used in this study therefore required an extensive search across multiple repositories and careful manual curation, combining the Zenodo MDMFR dataset (Ullah and Javed, 2022), the Kaggle CelebA-HQ masked collection (Dubey and Pawar, 2024), and real-world images collected directly by the author. The construction of this paired dataset is, in consequence, itself part of the contribution of the present study. Expanding it in the number of identities and in the number of paired images per identity is a continuing effort, and as larger paired collections are assembled or become publicly available, the strong separability reported in Section 5 can be re-examined at correspondingly larger scale.

A second, closely related difficulty concerns occlusion diversity within identities, and it directly explains the behavior of the OACT occlusion score analyzed in Section 5.6. If paired masked-unmasked images of the same person are scarce, then images of the same person under several different occluders a scarf, a surgical or cloth mask, an N95 respirator, a transparent face shield, or eyewear such as sunglasses that intrudes on the periocular zone itself are scarcer still, and obtaining them at controlled, graded levels of coverage is harder again. Considerable effort was made to include varied occlusion conditions in the dataset, and Figures 10 and 11 show the diversity that was achieved across identities, spanning surgical masks, cloth coverings, and an N95 respirator. Within any single identity, however, most subjects could only be obtained under one predominant mask type, so the jointly trained occlusion head observes little coverage variation per person and, supervised only by a binary mask-presence label, learns the narrow 0.43-0.52 score band reported in Section 5.6, which in turn limits the per-probe OACT adjustment to 3-4% on this dataset. The mechanism itself trains stably, integrates cleanly into the verification pipeline, and adds no inference cost; realizing its full adaptive range is therefore a data-collection problem before it is a modelling one. Future work will extend the dataset with multiple occluder types and coverage levels per identity and train the occlusion head against graded mask-coverage labels, allowing OACT to exert the larger per-probe adjustments it was designed for.

These two limitations define a single, coherent direction for future work: growing the paired dataset in both identity count and per-identity occlusion diversity, and pairing that growth with graded occlusion supervision. Both extensions build directly on the data-collection methodology established in Section 3, and neither alters the findings already obtained: on the data that genuinely paired collections currently allow, anatomically guided periocular attention combined with global Transformer modelling produces a discriminative, interpretable cross-modal embedding.

## 7. Conclusion

This paper presented PLGSA-Transformer, a cross-modal face matching framework for recognizing individuals across masked and unmasked conditions. The framework combines three components with deliberately distinguished scope, as set out in Section 1: Periocular Landmark-Guided Spatial Attention (PLGSA), a new mechanism that injects an explicit anatomical prior derived from MediaPipe facial landmarks directly into the spatial attention computation through a learnable residual gate; a Hybrid CNN-Transformer Branch, an integration of EfficientNetB3 local feature extraction with global cross-regional dependency modelling through a Multi-Head Self-Attention Transformer encoder; and the Occlusion-Adaptive Cosine Threshold (OACT), a new per-probe threshold formula that is architecturally validated in the present experiments and whose full adaptive range awaits the richer occlusion diversity and graded supervision outlined in Section 6.

Evaluated on a dataset of 858 paired images assembled from three independent sources under an identity-disjoint protocol, the proposed model achieved a pair verification accuracy of 97.2% with a ROC AUC of 1.0000, exceeding the same-author VGG-16-based MUFM model and two traditional feature-based baselines; the paired dataset underlying this evaluation was assembled specifically for this study, reflecting the scarcity of same-person masked-unmasked data discussed in Sections 3.1 and 6. On the more demanding open-set gallery-probe identification task, the model achieved a Rank-1 accuracy of 78.0% and a Rank-5 accuracy of 96.0%, with analysis confirming that the learned periocular embedding clusters identities into tight neighborhoods such that the majority of Rank-1 errors place the correct identity at Rank-2 or Rank-3.

The results provide consistent and interpretable evidence that explicitly encoding periocular landmark geometry into the attention mechanism, combined with global Transformer modelling, is an effective and theoretically grounded approach to cross-modal masked face recognition. The complete absence of false acceptances on the verification task, the tight identity clustering revealed by the Rank-1/Rank-5 analysis, and the correct anatomical localization of the PLGSA attention maps across all qualitative cases together indicate that the proposed embedding captures genuinely discriminative periocular structure. Future work will proceed along the roadmap set out in Section 6: expanding the paired dataset in both identity count and per-identity occlusion diversity, and introducing graded occlusion supervision to realize the full adaptive potential of the OACT mechanism, with evaluation extended to larger paired benchmarks as they become available.